\documentclass{article}

\usepackage{subfigure}
\usepackage{wrapfig}
\usepackage{lipsum}
\usepackage[preprint]{neurips_2023}

\usepackage[utf8]{inputenc} % allow utf-8 input
\usepackage[T1]{fontenc}    % use 8-bit T1 fonts
\usepackage{hyperref}       % hyperlinks
\usepackage{url}            % simple URL typesetting
\usepackage{booktabs}       % professional-quality tables
\usepackage{amsfonts}       % blackboard math symbols
\usepackage{nicefrac}       % compact symbols for 1/2, etc.
\usepackage{microtype}      % microtypography
\usepackage[dvipsnames]{xcolor}

\usepackage{microtype}
\usepackage{graphicx}
\usepackage{subfigure}
\usepackage{booktabs} % for professional tables

\hypersetup{
    colorlinks=true,
    linkcolor=teal,
    filecolor=magenta,      
    urlcolor=ForestGreen,
    citecolor=NavyBlue,%BlueViolet,%teal,
    pdftitle={Bring Your Own Data! Self-Supervised Evaluation of Large Language Models},
    pdfpagemode=FullScreen,
    }

\usepackage{amsmath}
\usepackage{amssymb}
\usepackage{mathtools}
\usepackage{amsthm}
\usepackage{graphbox}
\usepackage{xcolor}
\usepackage[most]{tcolorbox}
\usepackage{soulpos}
\usepackage{multirow}

\usepackage[capitalize,noabbrev,nameinlink]{cleveref}

\usepackage[font=small,labelfont=bf]{caption}

\title{Bring Your Own Data!  Self-Supervised Evaluation of Large Language Models} 
\newcommand{\expnumber}[2]{{#1}\mathrm{e}{#2}}
\DeclareMathOperator{\ppl}{ppl}
\DeclareMathOperator{\JSD}{JSD}
\newcommand\blfootnote[1]{%
  \begingroup
  \renewcommand\thefootnote{}\footnote{#1}%
  \addtocounter{footnote}{-1}%
  \endgroup
}
\author{%
  Neel Jain* 
  \And
  Khalid Saifullah*
  \And 
  Yuxin Wen
  \And 
  John Kirchenbauer
  \And 
  Manli Shu
  \And 
  Aniruddha Saha
  \And Micah Goldblum $^{\dagger}$
  \And Jonas Geiping 
  \And  Tom Goldstein \And \\
  {University of Maryland \quad $^{\dagger}$New York University  
    }
}

\begin{document}

\maketitle

\begin{abstract}
\looseness -1 With the rise of Large Language Models (LLMs) and their ubiquitous deployment in diverse domains, measuring language model behavior on realistic data is imperative. For example, a company deploying a client-facing chatbot must ensure that the model will not respond to client requests with profanity. Current evaluations approach this problem using small, domain-specific datasets with human-curated labels. These evaluation sets are often sampled from a narrow and simplified distribution, and data sources can unknowingly be leaked into the training set, which can lead to misleading evaluations. To bypass these drawbacks, we propose a framework for {\em self-supervised evaluation} of LLMs by analyzing their sensitivity or invariance to transformations on the input text. Self-supervised evaluation can directly monitor LLM behavior on datasets collected in the wild or streamed during live model deployment. We demonstrate self-supervised evaluation strategies for measuring closed-book knowledge, toxicity, and long-range context dependence, in addition to sensitivity to grammatical structure and tokenization errors. When comparisons to similar human-labeled benchmarks are available, we find strong correlations between self-supervised and human-supervised evaluations. The self-supervised paradigm complements current evaluation strategies that rely on labeled data. Code is available at \url{https://github.com/neelsjain/BYOD}. \blfootnote{* Equal contribution. Correspondence to: Neel Jain <njain17@umd.edu>.}

\end{abstract}

\section{Introduction}
As Large Language Models (LLMs) continue to advance rapidly, there has been a growing demand for new evaluation metrics that can accurately capture their capabilities and limitations \citep{ethayarajh2020utility, birhane2022values, kiela2021dynabench, bowman2021will}. As a result, there has been a constant need to create new datasets as newer models continuously make the existing datasets obsolete. Recent approaches such as BIG-Bench \citep{srivastava2022beyond} and HELM \citep{liang2022holistic} aim to address this issue by providing an ever-increasing, diverse set of accumulating micro-benchmarks to measure the performance of LLMs. However, these approaches still rely heavily on dataset creation and curation, which is time-consuming and expensive. 

Furthermore, evaluation is generally \textit{dataset-centric}, meaning that evaluations are based on some human-labeled or generated metric evaluated on a fixed dataset. For modern LLMs, this conventional approach comes with new complications. First, evaluation data is hosted on the internet (for example on sites like GitHub). This makes them accessible to scraping bots that generate training data for LLMs, making older datasets unreliable unless they are painstakingly removed from the training set, which does not reliably happen \citep{brown2020language, eval-harness}.\footnote{Efforts such as \url{https://github.com/hitz-zentroa/lm-contamination} are trying to catalog this phenomenon for ChatGPT.} Second, LLM evaluation is by its nature multi-faceted, since different LLM applications rely on distinct capabilities, and an ever-increasing number of such capabilities needs to be tested in modern LLMs. As dataset curation is expensive, each test in a large benchmark like HELM \citep{liang2022holistic}, uses only a small dataset -- carefully created to test a particular capability in a particular scenario. However, models are then deployed in much broader contexts and settings, and the applicability of these evaluations to deployment usage can be uncertain. 

\looseness -1 To complement conventional evaluation, we propose a framework for \textit{self-supervised model evaluation}. In this framework, metrics are defined as invariances and sensitivities that can be checked in a self-supervised fashion using interventions based only on the model in question rather than external labels. Self-supervised evaluation pipelines are \textit{dataset-agnostic}, and so they can be utilized over larger corpora of evaluation data than conventional metrics, or even directly in production systems to monitor day-to-day performance. In this work, we develop this framework, discuss desiderata for such metrics, and provide several case studies for self-supervised metrics: measuring knowledge through negations, toxicity detection, long-range dependency, word-order, and tokenization sensitivity. 
By developing these new metrics, we hope to provide a more comprehensive and nuanced understanding of the strengths and limitations of LLMs. 

\begin{figure}[t]
    \centering
    \includegraphics[width=\textwidth]{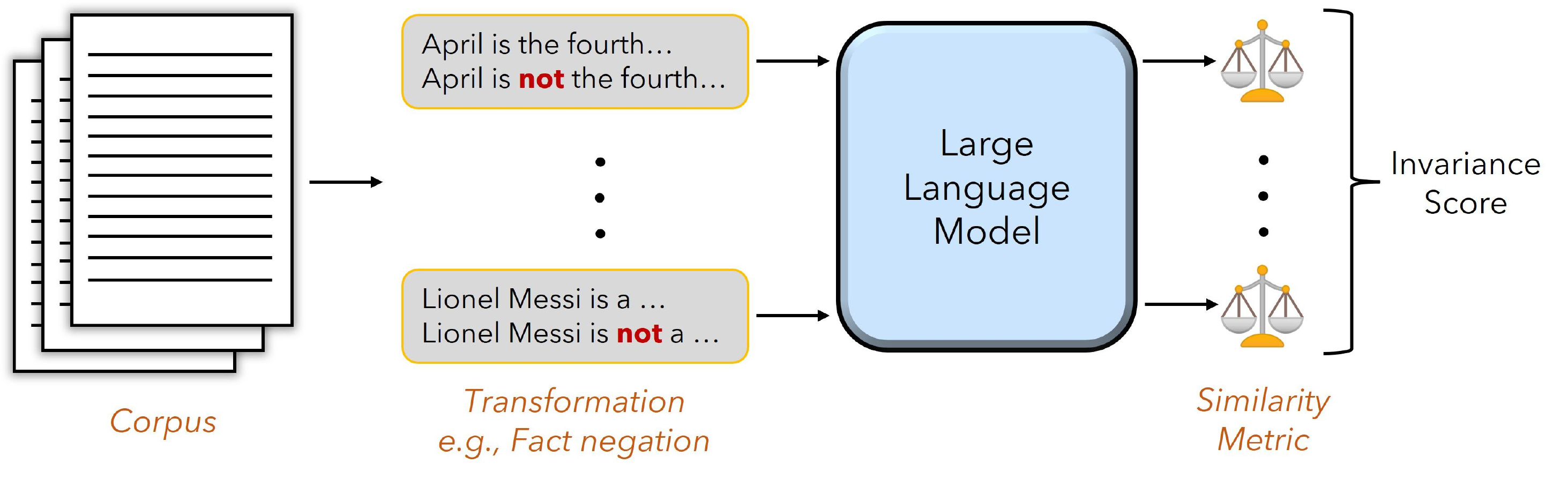}
    \caption{In our proposed self-supervised evaluation, pairs are created from a corpus. Each pair contains the original and perturbed text, which in the figure above is creating a negation via applying a ``not.'' These pairs are then fed into the network, and the outputs (perplexity, probability distributions, or text) are compared for each pair. These measures are then aggregated to produce an invariance or sensitivity score.}
    \label{fig:teaser}
    \vspace{-.3cm}
\end{figure}

\section{A Procedure for Self-Supervised Evaluation}

Our goal is to measure properties of LLMs such as toxicity, closed-book knowledge, and word order sensitivity without relying on benchmark-specific datasets or human annotations.  Rather than measuring model accuracy against known ground truth labels, we choose a simple transformation that can be applied to text.  We then measure the level of invariance that a model's output has under that transformation.  If we choose our transformations carefully, we can obtain useful information about model behavior in a completely self-supervised way.

\begin{figure}[t]
    \centering
    \includegraphics[width=0.45\linewidth]{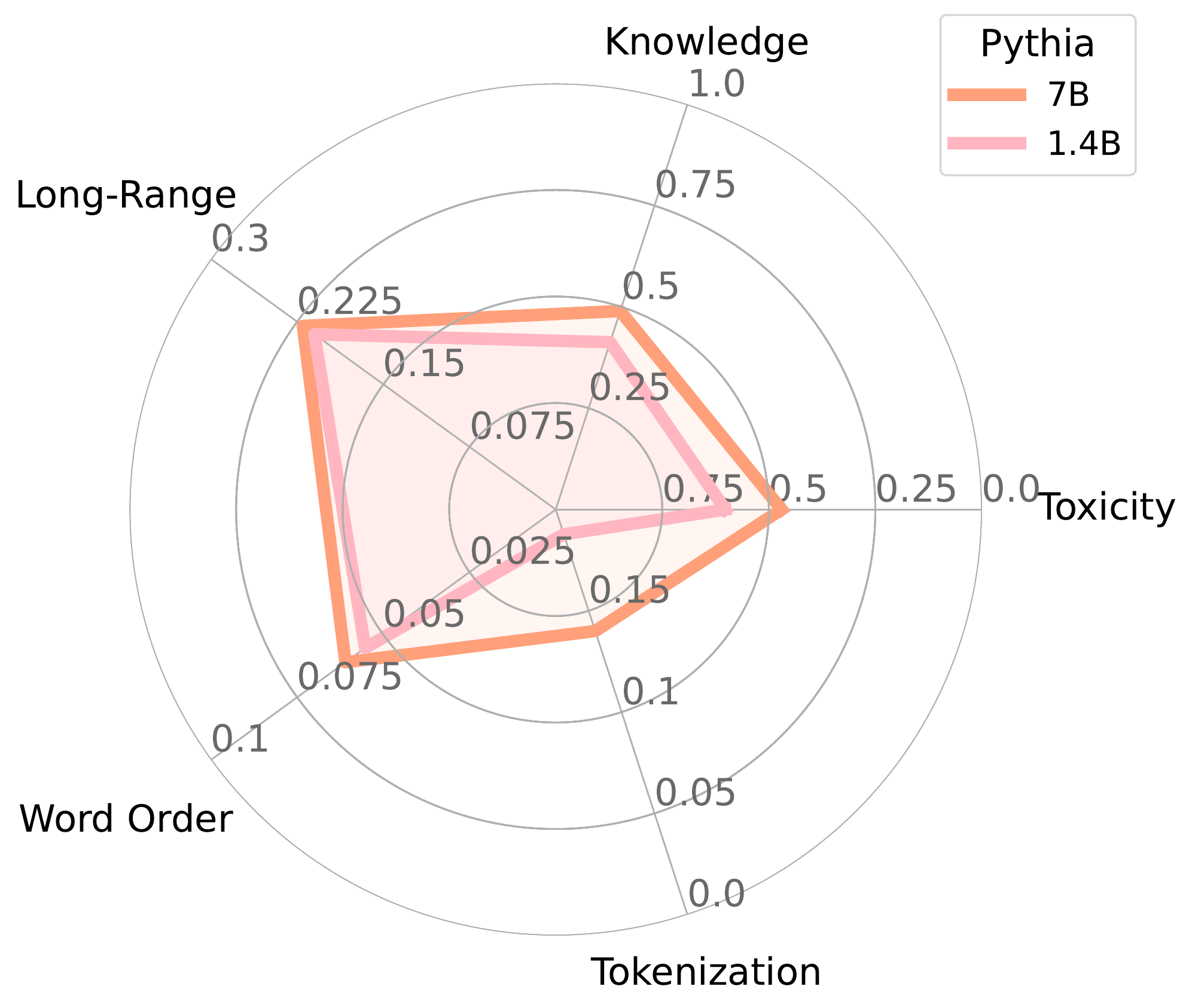}
    \includegraphics[width=0.45\linewidth]{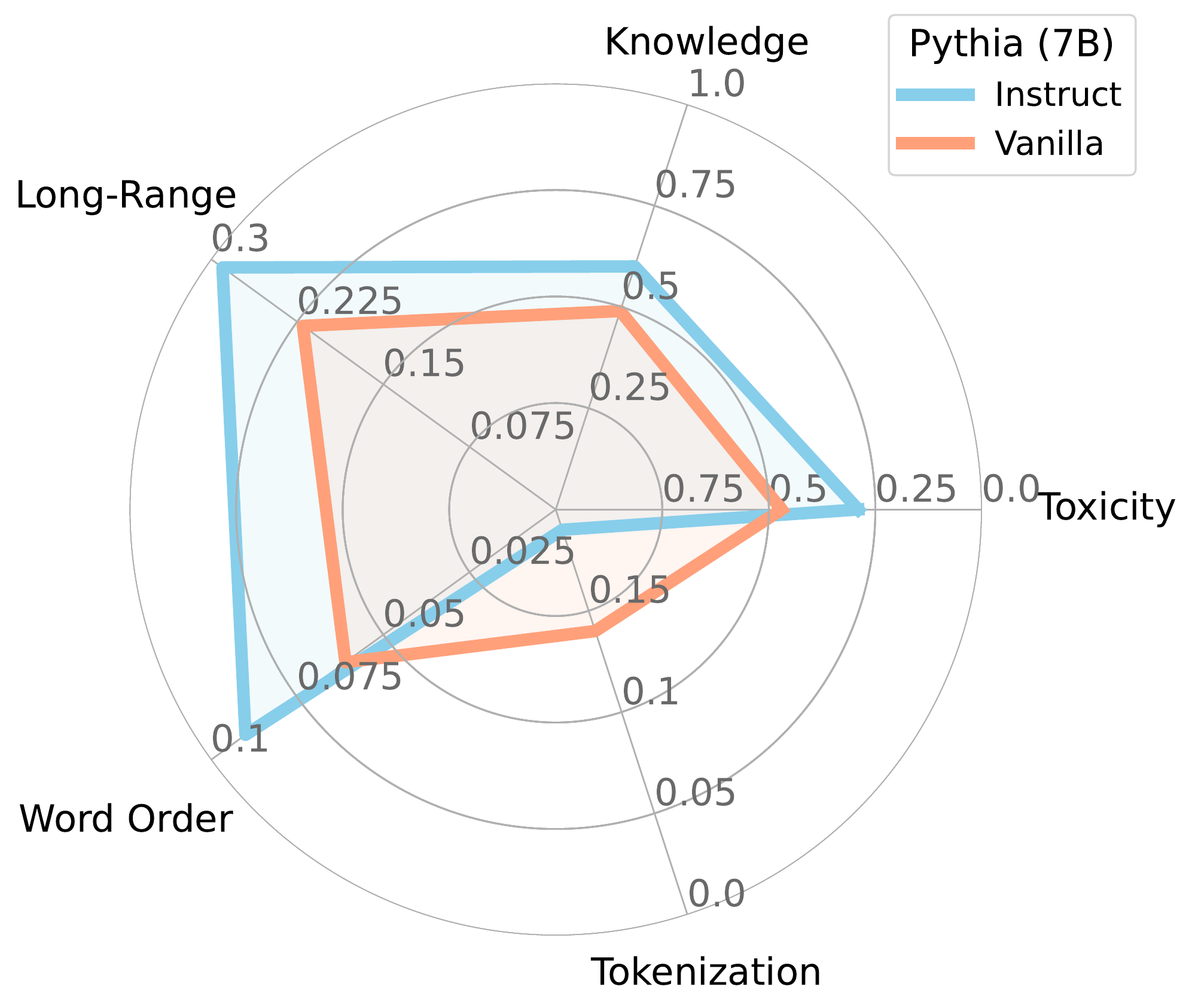}
    \caption{Spider plots showing sensitivity scores for the \emph{Knowledge Probing via Negations}, \emph{Toxicity}, \emph{Context (Long-Range)}, \emph{Word Order}, and \emph{Tokenization} metrics introduced in the paper. A larger area corresponds to a better model in terms of the sensitivity scores. \textbf{(Left)} Comparison between Pythia 1.4B and Pythia 7B models. The larger model performs better for all the metrics. \textbf{(Right)} Comparison between the instruction finetuned version (Dolly-V2) and the vanilla Pythia model. The instruction finetuned model is better than the vanilla model for all metrics except tokenization robustness.}
    \label{fig:pythia_spiderplots}
\end{figure}

More concretely, given a corpus $D$ (e.g., Wikipedia), we construct pairs of original passages/sentences $x$, and transformed counterparts $x'$. An example is seen in \cref{fig:teaser}, where we negate the original sentence $x$ to construct $x'$. $X$ is the set of all transformed pairs. We then feed input pairs into the language model, $f$, to extract a pair of outputs. Depending on the construction, the output being considered can be the softmax probability vector over tokens, a perplexity score, or a feature vector. We then compare the outputs $f(x)$ and $f(x')$ using a similarity metric, $\mathcal{M}$. Finally, we aggregate the results over all pairs in the data corpus using an aggregation operator, $A$, to produce an invariance/sensitivity score.
\begin{equation}
    \textsc{score} = A\{\mathcal{M}(f(x),f(x'))\text{ } \forall (x,x')\in X \}.
\end{equation}
In this work, we bring \texttt{wikipedia} as our own dataset, but note that we do so to enable comparisons to existing metrics that use human labels on similar data. We use this methodology to study several case studies, namely knowledge via negations (\cref{sec:knowledge}), toxicity (\cref{sec:toxicity}), context sensitivity (\cref{sec:context}), word order sensitivity (\cref{sec:word_order}), and tokenization robustness (\cref{sec:tokenization}) culminating in sensitivity scores as seen in \cref{fig:pythia_spiderplots}. In practice, these metrics should not be constrained to this data source, but evaluated directly on application-relevant sources.
\section{Related Work}

HELM adopts a multi-metric approach: accuracy, calibration, robustness, fairness, bias, toxicity, and efficiency over each of the datasets proposed \citep{liang2022holistic}. These metrics build on the work of \citet{ribeiro-etal-2020-beyond} and subsequent studies such as, \citep{Mille2021AutomaticCO, Wu2021PolyjuiceGC, Ross2021TailorGA, dhole2021nl, Yang2022TestAugAF} which augment inputs from a dataset to measure properties beyond the classical metric of accuracy. While these methods rely on existing datasets and labels, our method departs from these previous works as we analyze invariances using a data-agnostic procedure.

\looseness -1 \textbf{Knowledge Probing via Negation:} 
The MMLU benchmark \citep{hendrycks2021measuring} is widely used to assess the knowledge base of language models, evaluating their performance on task-specific micro datasets. In production, the GPT-4 technical report \citep{openai2023gpt4} advertises the model's capabilities across various knowledge categories, yet the evaluation suite used in the report is not publicly available.  Furthermore, \citet{Wu2021PolyjuiceGC} introduces a general-purpose counterfactual generator, \textit{Polyjuice}, that allows for control over perturbation types and locations and is trained by finetuning GPT-2 on multiple labeled datasets of paired sentences. In contrast, we focus on evaluating the knowledge base of LLMs through invariances where no labeled data is required. 
\textit{Negations:} \citet{ettinger2020bert} utilize psycholinguistic tests to explore the general linguistic knowledge and contextual impacts of negation in language models.
Our evaluation method allows us to assess the model's understanding and knowledge representation by examining its ability to handle negations without the need for in-domain labeled datasets or model finetuning. 

\textbf{Toxicity:} \texttt{RealToxicityPrompts} is the most prominent benchmark for toxicity in LLMs \citep{gehman_realtoxicityprompts_2020}. This method relies on the \texttt{Perspective API}\footnote{\url{https://perspectiveapi.com/}} to score the model's generation based on a series of prompts. This API is also used as the toxicity metric for HELM. However, with the proprietary API constantly changing, comparing evaluations across time is difficult \citep{pozzobon2023challenges}. Another common benchmark is \texttt{BOLD}  \citep{BOLD}. \texttt{BOLD} trains another model to classify toxic generations.
This approach of utilizing another model to measure toxicity is common \citep{sun-etal-2022-safety}.
Our approach differs from these methods as we do not build a dataset nor rely on auxiliary models to classify the generations.

\textbf{Word Order:} 
While previous efforts have made significant contributions to testing the compositional and word order understanding of language models \citep{oconnor2021context, thrush2022winoground}, these efforts predominantly rely on small sets of hand-crafted examples. Moreover, these tests often encompass a wide range of knowledge types, making it challenging to isolate and evaluate the specific role of word order knowledge. Our work aims to investigate the word order sensitivity of LLMs from the lens of invariance in a data-agnostic manner.

\textbf{Long-Range Dependency:} 
As conversational AI models become more prevalent \citep{ouyang2022training, Anthropic_Claude_2023}, the importance of accommodating large context lengths has become evident. Recent endeavors have focused on developing chat models with extensive context capabilities, such as 32k and 100k \citep{openai2023gpt4, Anthropic_2023}, utilizing techniques like memory-efficient attention \citep{dao2022flashattention}. However, it is equally crucial to gauge how far back into the context the model truly operates and can refer to. 
LAMBADA \citep{paperno2016lambada}, addresses this by assessing language models' comprehension of broad contexts.  
In contrast, our self-supervised approach creates texts through closed-form transformations that evaluate language models' grasp of long-range sensitivity.

\textbf{Tokenization Sensitivity:} HELM approaches this problem by inducing spaces, misspellings, etc. over the datasets in question to determine if these slight changes can affect changes when evaluating over established datasets \citep{liang2022holistic}. Additionally, \citet{MagikarpToken} found a set of anomalous tokens which result in a previously undocumented failure mode for GPT-2 and GPT-3 models. Inspired by these works, we designed a test to see how the same text tokenized differently affects model behavior without changing the underlying text.

\section{Knowledge Probing via Negations: \texttt{Au Contraire Metric}} \label{sec:knowledge}

This section presents a simple self-supervised evaluation for knowledge probing.
Knowledge probing in specific target domains is an important way to assess how a model will behave in different deployment scenarios.
OpenAI approached this problem by constructing nine adversarial datasets on varying areas such as Law and Technology to evaluate GPT-4 \citep{openai2023gpt4}. While OpenAI's approach and others like MMLU \citep{hendrycks2021measuring} are a step forward, these datasets do not cover all possible domain-specific areas. Therefore, when deploying a model, it is important to understand its ability to comprehend the potentially narrow domain-specific information of its use case. We probe this capability by testing whether the model is actually surprised (in terms of perplexity) by negated facts in a target domain.

\textbf{Self-Supervised Approach:} We construct a simple transformation over factual information like definitions by applying negations to facts. This is done in a trivial self-supervised way:  We search for the first occurrence of \texttt{is}, \texttt{was}, or \texttt{were}, and place the word \texttt{not} after it provided a negation is not already present. For example, given the fact ``\texttt{April is the fourth month of the year in the Julian and Gregorian calendars and comes between March and May}'', we apply the negation transformation to this sentence and construct: ``\texttt{April is \textit{not} the fourth month of the year in the Julian and Gregorian calendars and comes between March and May}''. 

Based on this intervention, we measure the change in the log-perplexity ($\log(\ppl(x))$), between the original and negated sentence. Formally, we define the sensitivity score as the following:
\begin{equation*}
    \textsc{sensitivity score} = \frac{1}{n}\sum^n_i \log (\ppl(x_i')) - \log(\ppl(x_i)).
\end{equation*}
One possible confounding variable is how sensitive a model is to the term ``not'' in a sentence. 
One way to normalize this behavior is to approximately measure the model's sensitivity to ``not'' over a benign corpus, where the meaning of ``not'' should not have a sizable impact on the perplexity over sentences nor have a known expected direction: 
\begin{equation*}
    \textsc{normalized sensitivity score} = \textnormal{sensitivity score} -  \frac{1}{m}\sum^m_i |\log (\ppl(y_i')) - \log(\ppl(y_i))|,
\end{equation*}
where $y$ is a sample from a benign corpus like \texttt{bookcorpus} with $m$ total samples for which there is not a clearly defined truth value. Note that we use the absolute value of the difference, as it is unclear which direction is expected from the model for a given input in the benign corpus. To evaluate the relationship of these metrics to model confidence in our analysis, we also record the fraction of inputs for which perplexity decreases after introducing a negation, which represents, for a typical sample, the error that the model is making: $\textsc{Percent PPL Drops} = \frac{1}{n}\sum^n_i \max{\{\text{sign}(\log (\ppl(x_i)) - \log(\ppl(x_i'))),0\}}.$

\subsection{Experimental Set-up}
To verify that this self-supervised evaluation is sensible, we compare our method to accuracy on TriviaQA, as both evaluations gauge an LLM's world knowledge \citep{joshi2017triviaqa}. We do not penalize the length of the output. More details on exactly how we calculate accuracy can be found in the Appendix. Since TriviaQA asks general knowledge questions, we apply our self-supervised metric to topic sentences from Wikipedia to get a comparable general knowledge score. 
A human inspection of 100 samples verified that the proposed transformation resulted in grammatically correct sentences that were counterfactuals for the original sentence.
To calculate our metric, we measure the sensitivity score over $1000$ examples, where the standard error for these scores was less than $0.002$. Since we use perplexity, we can also utilize API models, such as those from OpenAI and Cohere, and publicly available models from the Hugging Face Hub, such as Pythia and GPT-2 \citep{biderman2023pythia, brown2020language, radford2019language}. A full list of the models we evaluate can be found in the Appendix. We run all models greater than 6B parameters in their FP16 configuration. 

\begin{figure*}[h!]
    \centering
    \includegraphics[width=0.365\linewidth]{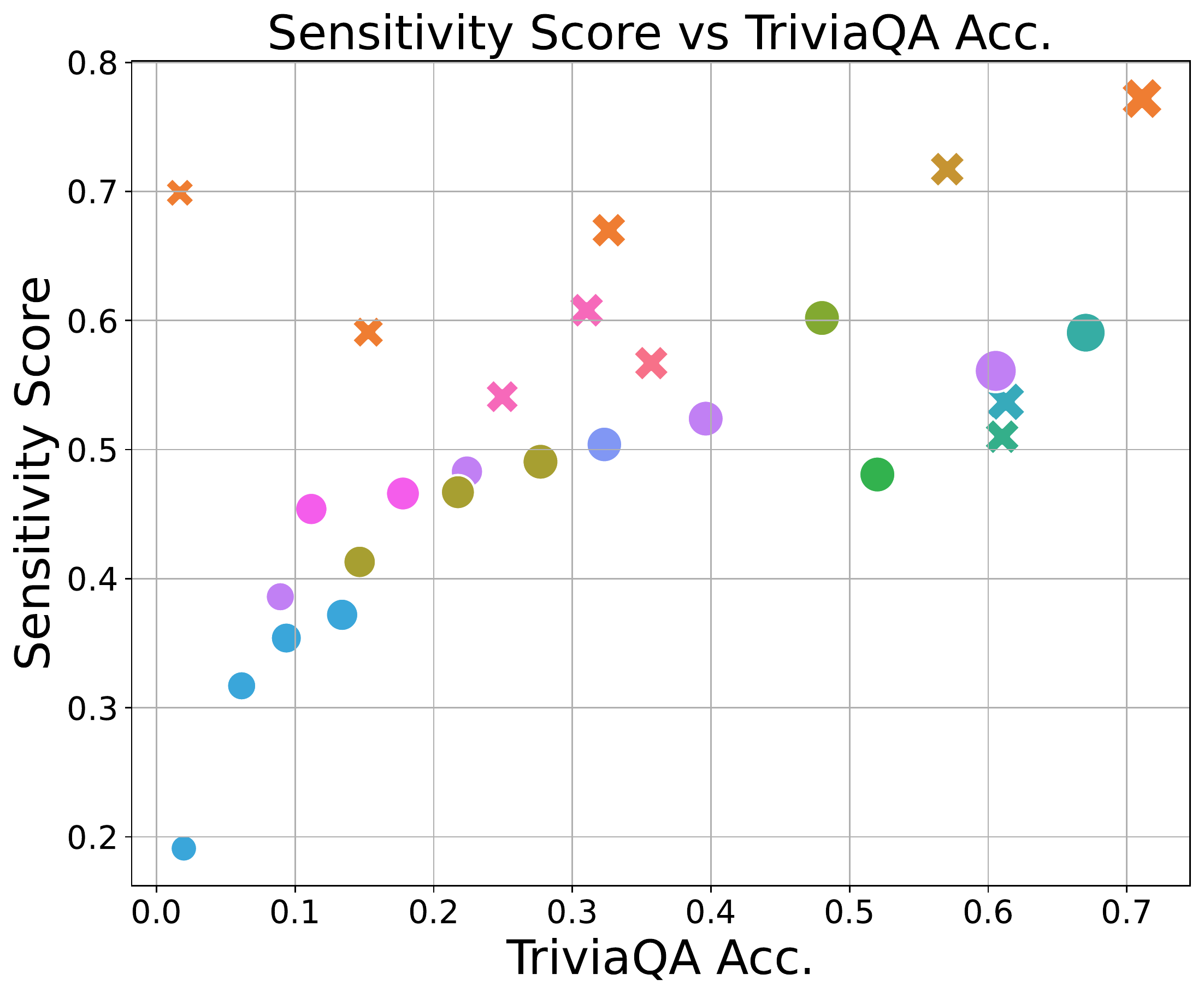}
    \includegraphics[width=0.5\linewidth]{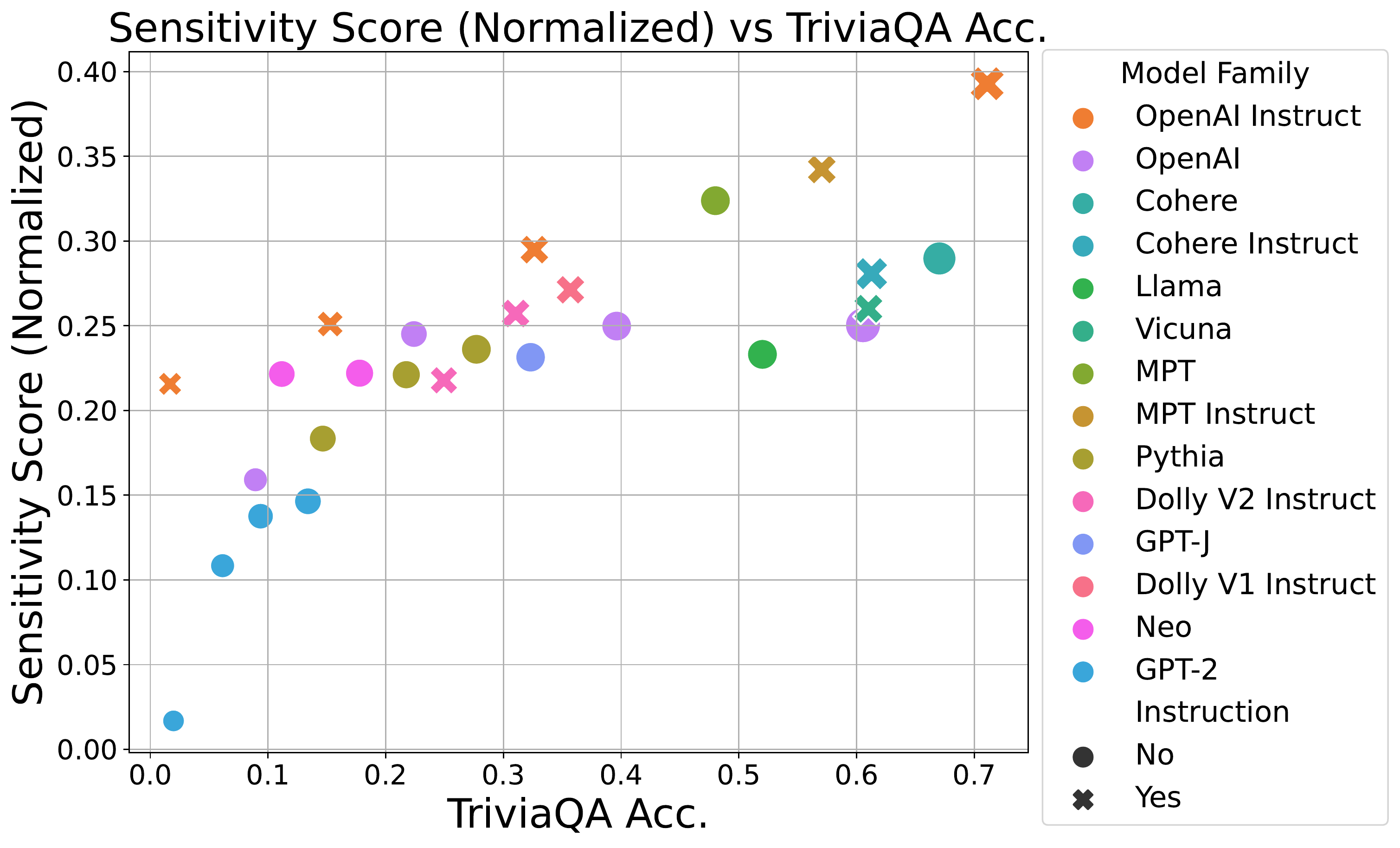}
    \caption{\textbf{(Left)} Sensitivity Score (negations) compared to accuracy on TriviaQA over various model sizes and families. \textbf{(Right)} Normalized Sensitivity Score compared to accuracy on TriviaQA over various model sizes and families. Larger markers correspond to bigger models, and ``x'' markers represent instruction finetuned models.}
    \label{fig:Negations_All_Models_both_plots}
    \vspace{-.4cm}
\end{figure*}
\subsection{Results}
\begin{wrapfigure}[16]{r}{0.5\textwidth}
    \vspace{-0.45cm}
    \includegraphics[width=0.5\textwidth]{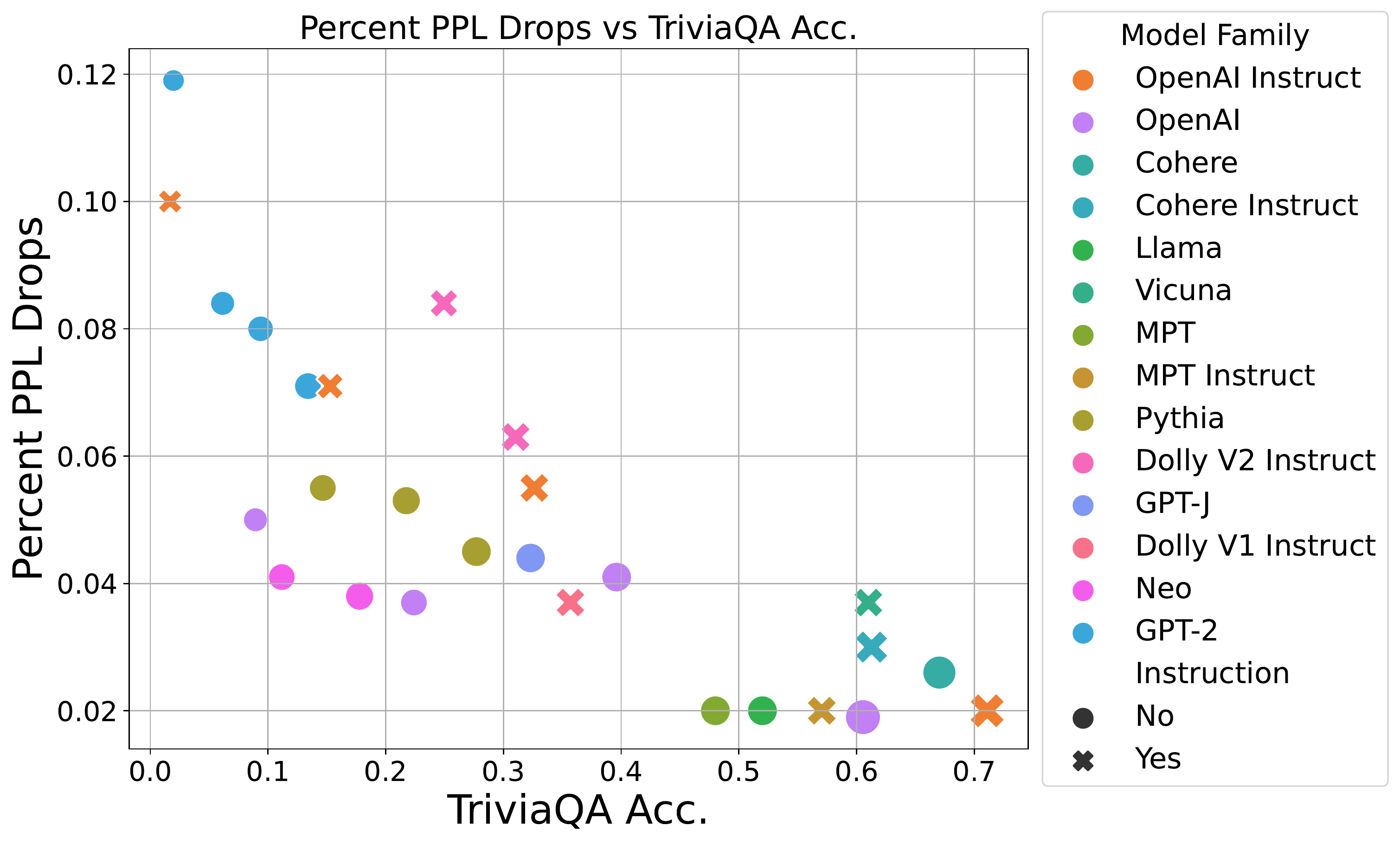}
    \caption{Percentage of samples where perplexity drops versus accuracy on TriviaQA. We observe a reliable negative correlation.}
    \label{fig:Negations_Percent_Wrong_Direction}
    \vspace{-.25cm}
\end{wrapfigure}
\cref{fig:Negations_All_Models_both_plots} shows that the self-supervised \textsc{Sensitivity Score}, which measures the change in $\log(\ppl)$ over the pair of sentences, closely tracks accuracy on the human-curated TriviaQA dataset, especially for non-instruction finetuned models. It also maps closely to a square-root relationship, with normalization further improving this trend. Normalization corrects the instruction-tuned models to a larger degree, possibly due to their innate overconfidence. 
We can further hone in on why correct normalization is important by cross-referencing the frequency with which perplexity goes down rather than up, in \cref{fig:Negations_Percent_Wrong_Direction}. This ablation metric is robust to outlier perplexity values. Here, instruction-tuned models are well-behaved. 
Further, we notice that outliers in \cref{fig:Negations_All_Models_both_plots} are indicative of important model properties and weaknesses of the TriviaQA benchmark. For example, consider Cohere's instruction model (Cohere command), which has low sensitivity score relative to its TriviaQA performance and appears as a dark turquoise ``$\times$'' on the middle right of the chart, and text-ada-001 (OpenAI's smallest instruction model), which appears as an orange ``$\times$'' on the upper left side of the chart.  To investigate these outliers further, we applied negations to questions in TriviaQA and found that Cohere command model rarely changed its answer when a negation was introduced, whereas text-ada-001 changed its answer frequently. We show examples of this behavior in \cref{tab:negation_example_TriviaQA}.
This implies that the Cohere model is insensitive to sentence structure when the negation is present -- it has memorized the associations between concepts and answers based on the context alone, even if the construction of the question makes its answer incorrect. This inability to answer grammatically complex questions is not reflected in the TriviaQA results, due to its reliance on simple sentence structures and nearly uniform question formats. Text-ada-001 is the opposite, it is exceedingly sensitive to sentence structure and nearly always flips its answer when faced with a negation.  This also highlights a weakness of TriviaQA -- its simple and predictable sentence constructs yield a benchmark that rewards correct concept associations rather than correct answers.

In summary, we find that we can predict benchmark performance exceedingly well with a simple self-supervised scheme, validating the effectiveness of this metric.

\begin{figure}[b]
    \centering
    \vspace{-.2cm}
    \includegraphics[width=0.85\linewidth]{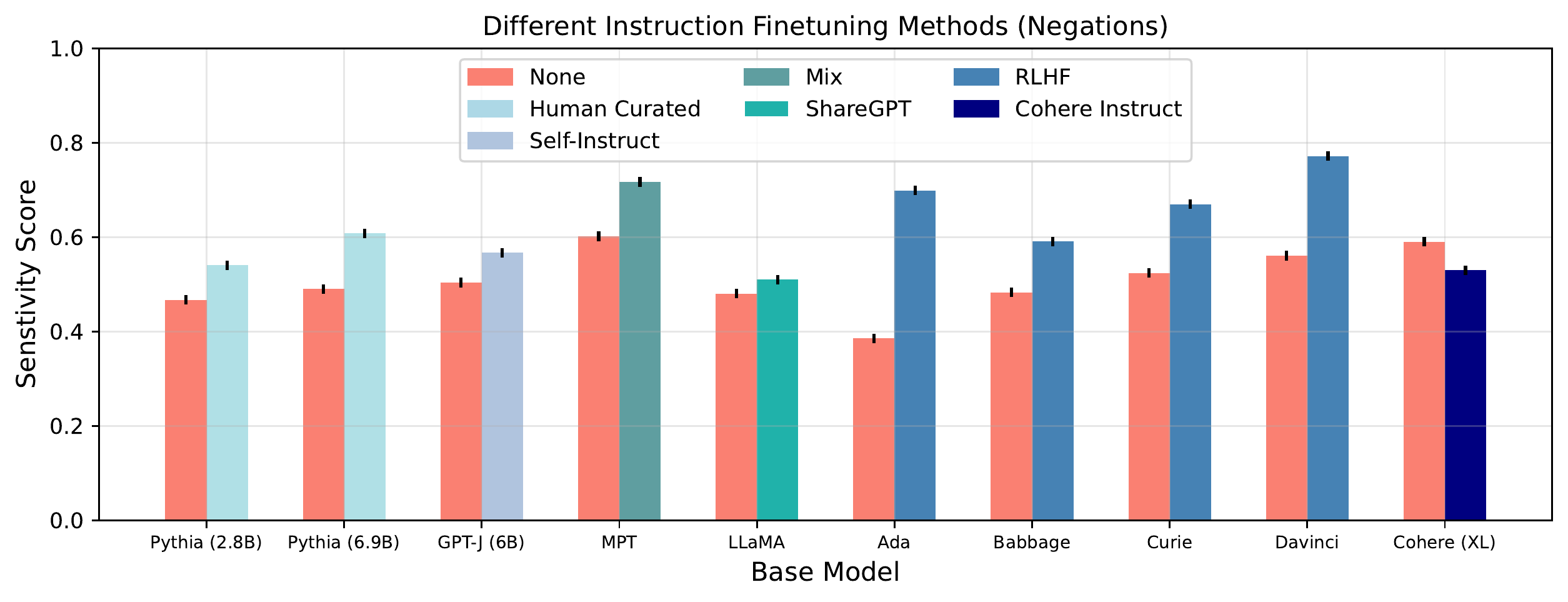}
    \caption{Sensitivity Score (negation) comparing pretrained LLMs with their instruction finetuned counterparts. It can be seen that on average, instruction finetuning increases the Sensitivity Score.}
    \label{fig:neg_instruct_tuned}
    \vspace{-.1cm}
\end{figure}

\textbf{Effect of Instruction Finetuning:} In general, we find that instruction-tuned models are more sensitive to negations than other LLMs as seen in \cref{fig:neg_instruct_tuned}, regardless of the source of instruction data. % This is surprising, given that 
% This may indicate that more nuanced things are occurring in instruction-tuned models in comparison with their vanilla counterparts. 
The outlier here is again the Cohere command model, which is less sensitive than Cohere's base model after finetuning.

\begin{table}
\small
\caption{Example outputs of text-ada-001, text-davinci-003 and Cohere command. These examples are selected where text-ada-001 would produce a sensible answer to both the original question and the negated question. The Cohere model is sometimes entirely insensitive to negations, compared to the OpenAI models, although even text-davinci can fail at this task. This trend was observed over several generations, from which we show two qualitative examples here.}
\begin{tabular}{lll}
\multicolumn{1}{c}{\textbf{Model}} & \multicolumn{1}{c}{\textbf{Original}}                                                                              & \multicolumn{1}{c}{\textbf{Transformed}}                                                                                                                                                                              \\ \toprule
   \textbf{Question}                                &\textbf{ A sterlet is what type of creature? }                                                                               & \textbf{A sterlet is \texttt{not} what type of creature?}                                                                                                                                                                               \\
\hline
text-ada-001                       & \begin{tabular}[c]{@{}l@{}}A sterlet is a creature that has \\ a spiny body and a long, sharp tongue.\end{tabular} & A sterlet is not a creature.                                                                                                                                                                                          \\
text-davinci-003                   & A sterlet is a type of sturgeon.                                                                                   & A sterlet is a type of sturgeon.                                                                                                                                                                                      \\
Cohere command               & Fish                                                                                                               & Fish                                                                                                                                                                                                                  \\ \midrule
 \textbf{Question}                                  & \begin{tabular}[c]{@{}l@{}}\textbf{What is the only natural food}\\ \textbf{ that never goes bad?}\end{tabular}                      & \begin{tabular}[c]{@{}l@{}}\textbf{What is \texttt{not} the only natural} \\ \textbf{food that never goes bad?}\end{tabular}                                                                                                                     \\
\hline
text-ada-001                       & \begin{tabular}[c]{@{}l@{}}The only natural food that \\ never goes bad is sugar.\end{tabular}                     & \begin{tabular}[c]{@{}l@{}}There is no one natural food that \\ never goes bad. There are, however, \\ some foods that are more likely to do so. \\ These include: milk, yogurt, ice cream,\\  and cake.\end{tabular} \\
text-davinci-003                   & Honey.                                                                                                             & \begin{tabular}[c]{@{}l@{}}There is no single natural food \\ that never goes bad.\end{tabular}                                                                                                                       \\
Cohere command                & Honey never goes bad.                                                                                              & Honey never goes bad.    \\\bottomrule                                                                                                                                                                                            
\end{tabular} \label{tab:negation_example_TriviaQA}
\end{table}

\textbf{Limitations:} For the sensitivity score to measure truthfulness, the dataset being used must contain a large fraction of sentences whose truth value is true, rather than neutral or false.  This is likely to hold for many corpora, if only to varying degrees. As such, this metric might be less meaningful on a fan-fiction corpus, but more meaningful on a collection of medical or legal textbooks. Finally, we chose a simple construction for the negation transform and found it to be effective.  LLMs like ChatGPT could be utilized to construct the counterfactual sentence when doing so would otherwise be non-trivial.  However, our simple construction has the benefit of reproducibility, as it does not depend on a commercial model that may change over time.

\section{Toxicity: \texttt{F-Bomb Metric}} \label{sec:toxicity}

Before LLMs are deployed in commercial settings, for example as a customer service chatbot, it is important to audit their potential to produce profanity or other toxic language.
Most methods for measuring toxicity involve feeding an LLM toxic prompts and then analyzing the outputs using a black-box commercial tool (e.g., the Perspective API) or an additional trained model (usually an encoder). However, using a model to measure the generation may be problematic. For example, although work like \citet{fortuna-etal-2020-toxic} has tried to understand how the \texttt{Perspective API} classifies toxic text, the API continues to change, and as it changes our understanding of how toxic generations are being classified starts to dissipate \citep{pozzobon2023challenges}.

\textbf{Self-Supervised Approach:} One simple and reproducible approach is to analyze toxic generation through invariance. We will construct a metric that quantifies how \textit{stoic} the model is to profanity, i.e., whether the model will respond to profane and aggressive comments with its own profanity or aggression. Although we study profanity, this can be extended to other forms of toxicity as well, or more broadly to model behaviors, such as tone, that the model should not mimic from user queries.
\begin{wrapfigure}[19]{r}{0.5\textwidth}
    \vspace{-.2cm}
    \includegraphics[width=\linewidth]{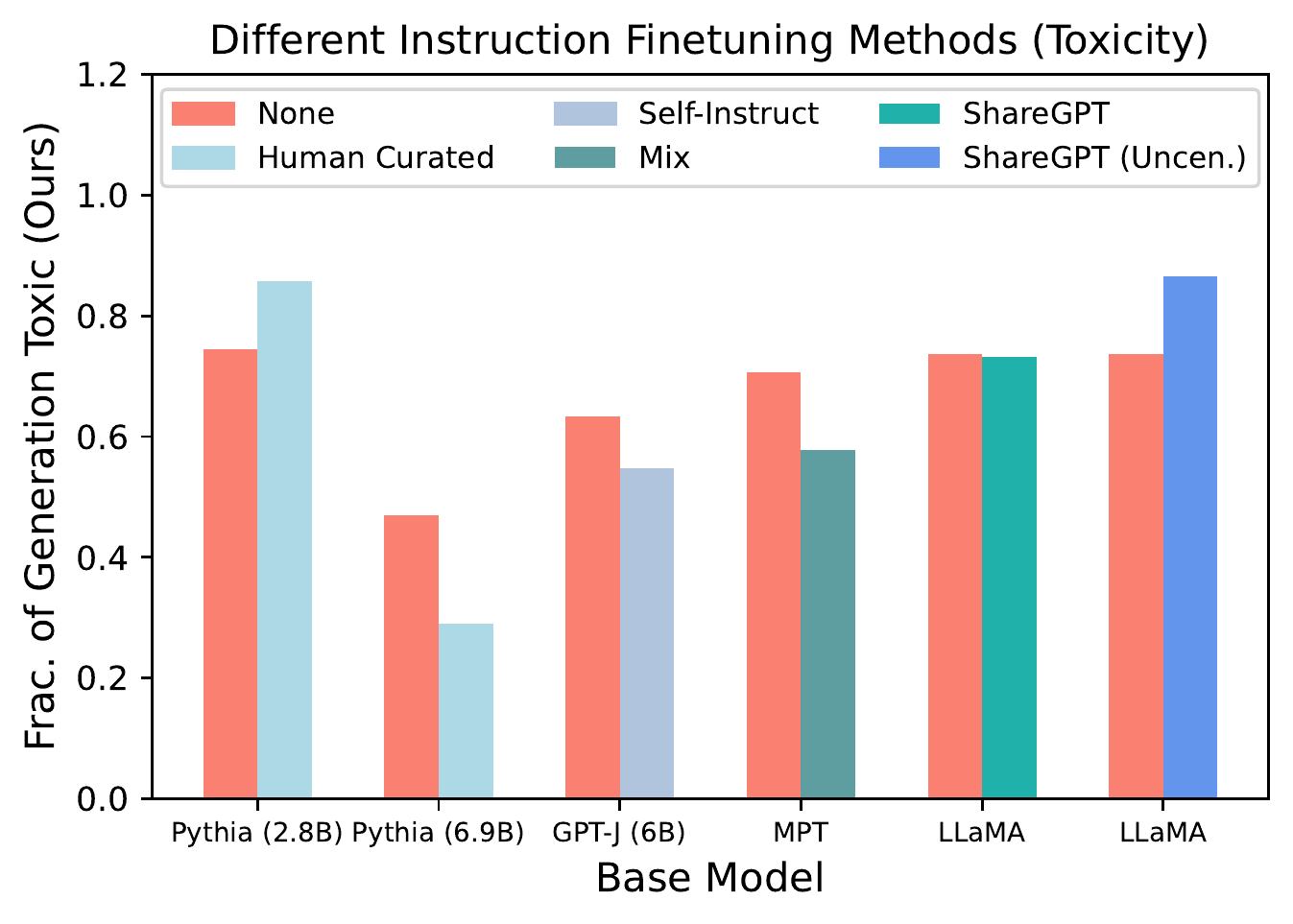}
    \caption{The fraction of generated outputs that are toxic (i.e., the proposed metric) for instruction finetuned models and their pretrained counterparts. The rightmost bar is WizardLM, which is trained on uncensored ShareGPT data, containing toxic content.}
    \label{fig:Toxicity_Instruction}
    
\end{wrapfigure}

To build a self-supervised evaluation, we transform each of our prompts by ``F-bombing:'' we append trigger words (e.g., ``F**k F**K'') at the end of the original text. We then feed these modified prompts into the model, examining the generated output. 

We construct two sensitivity scores: one that analyzes open-ended generated text, and another that analyzes the softmax probability distribution for the next predicted token. For the text generation score, we examine the output to determine if a profane word from the LDNOOBW dataset\footnote{\url{https://github.com/LDNOOBW/List-of-Dirty-Naughty-Obscene-and-Otherwise-Bad-Words/blob/master/en}} is contained in the generated text. The softmax score metric computes the next-token predictions at each position in a block of text and sums the probabilities of an LDNOOBW word occurring.
We validate that our measurement procedure is sensible by comparing our metric with \texttt{Perspective API} toxicity scores over the generations that contained the appended trigger phrase. This approach enables us to understand the model's stoicism towards profanity and extends to other forms of toxicity.

\subsection{Experimental Set-up}

We employ this self-supervised approach using the \texttt{wikipedia} dataset, which we split into prompts on a per-sentence level.  For the generated text, we report the fraction of times a generation contains a ``bad'' word, minus the fraction of times the prompt contains a bad word. 
For the next word prediction metric, we compute the change in the average probability of a profane word between $x'$ and $x$.
\begin{figure*}[t]
    \centering
    \includegraphics[width=0.378\linewidth]{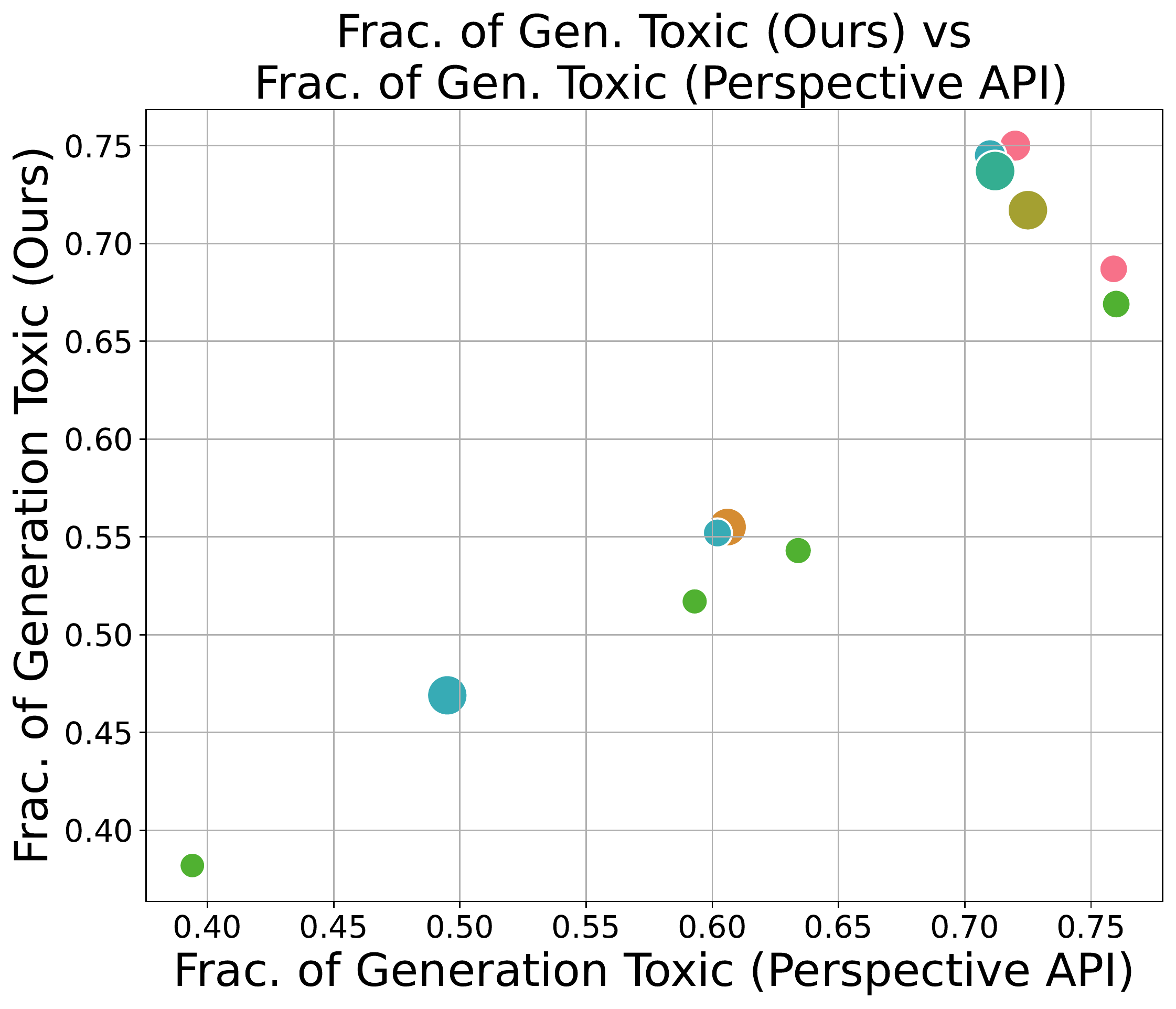}
    \includegraphics[width=0.48\linewidth]{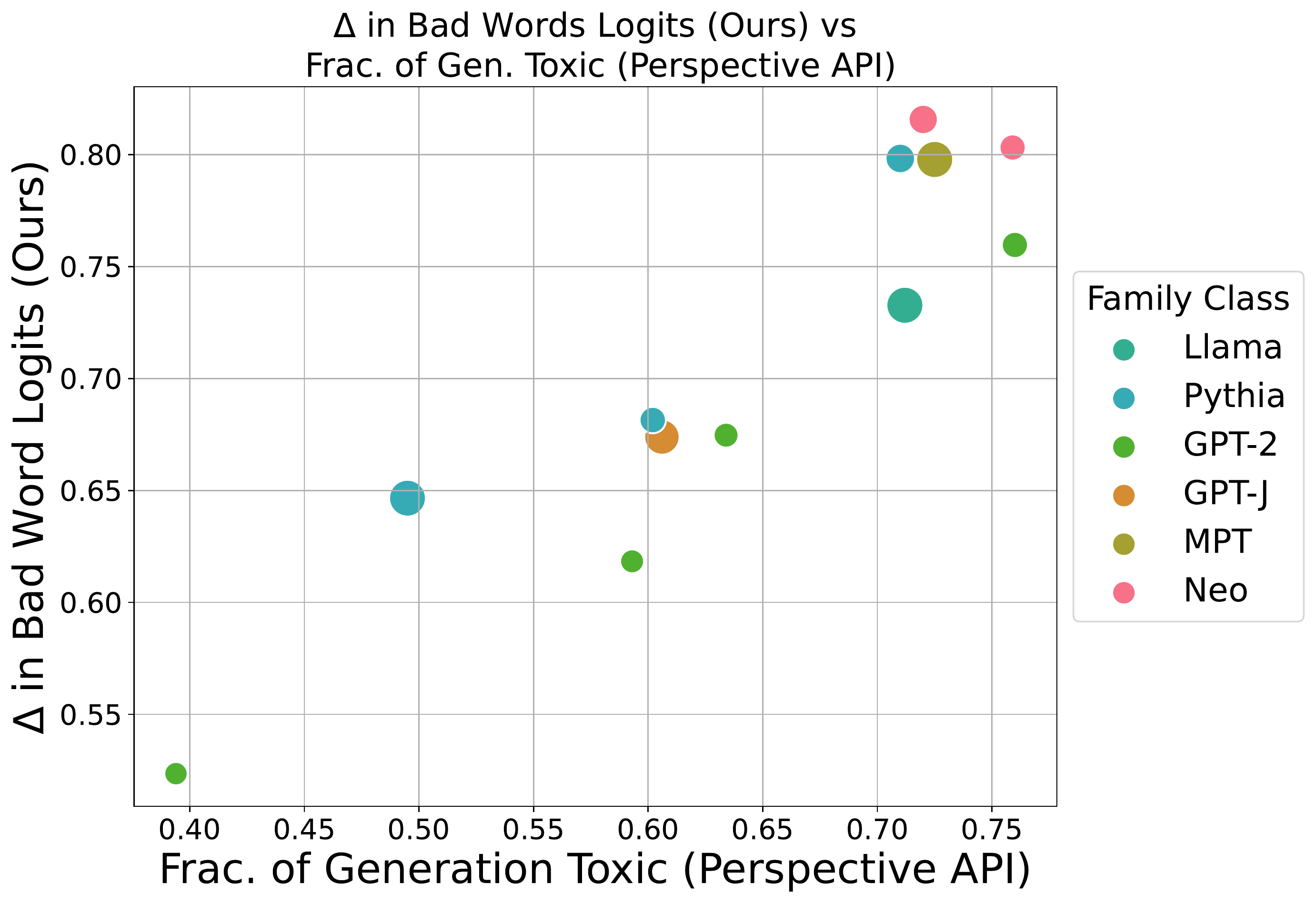}
    \caption{(\textbf{Left}) The fraction of toxic generations under our metric versus \texttt{Perspective API}. (\textbf{Right}) The change in the next-token probability of a profane word (token) versus the fraction of toxic generations according to \texttt{Perspective API}.}
    \label{fig:Toxicity_Logits_GenToxic}
    \vspace{-.5cm}
\end{figure*}
We compared our metric with the \texttt{Perspective API} \citep{fortuna-etal-2020-toxic} toxic scores on the prompts. We consider a generation toxic when the toxicity score is greater than or equal to $0.5$. 
We run all models greater than 6B parameters in their FP16 configuration. The full list of models can be found in the Appendix.

\subsection{Results}

The results of our toxicity metric, evaluated in both text space and logit space, are presented in \cref{fig:Toxicity_Logits_GenToxic}. The figures clearly demonstrate a close correlation between our metric, which measures the fraction of generated toxic word counts and the change in probabilities over the profane words, and the toxicity scores obtained from the Perspective API. We conducted tests using models of different types and scales (\cref{fig:Toxicity_Instruction} and \cref{fig:Toxicity_Logits_GenToxic}). Furthermore, from \cref{fig:Toxicity_Logits_GenToxic}, there appears to be no relation between the sensitivity of models to profane words and model size.

\textbf{Effect of Instruction Finetuning:} From \cref{fig:Toxicity_Instruction}, we see that seems to be no effect on average of instruction finetuning compared to their pretrained counterparts over the six models examined. The LLM with the lowest score is Dolly-V2 (7B), making it the least toxic model with respect to both our scores. Additionally, we see that MPT-Instruct is less toxic, which we suspect is due to the Harmless and Helpful dataset from Anthropic the model was trained on \citep{bai2022training}. Furthermore, we see that WizardLM, which is trained on an uncensored version of ShareGPT, is more toxic than a model trained on the filtered version of ShareGPT. While \citet{ouyang2022training} reported that RLHF decreases the toxicity of the model, this is ultimately highly dependent on the composition of the feedback data used to train the RLHF reward function. 

\textbf{Limitations:} Our analysis focuses on explicit profanity and may not capture nuanced forms of toxicity beyond explicit language. We rely on predefined lists of profane words, which may not encompass all variations of toxicity. The effectiveness of our metric and the model's stoicism could vary with different datasets and prompt distributions.

\vspace{-.1cm}
\section{Context (Long-Range) Sensitivity: \texttt{Back to the Future Metric}} \label{sec:context}
\vspace{-.1cm}

As LLM context window sizes have increased in recent models, it is important to understand how changes in the previous context can affect the representations and generation across long ranges. Datasets like Long-Range Arena \citep{taylong} offer a very broad set of tasks, focusing on context lengths ranging from $1k$ to, $16k$ and aim to evaluate architectural choices. There are other datasets like LAMBADA that focus on the capability to successfully predict the conclusion to a paragraph \citep{paperno2016lambada}. The dataset is designed such that the prediction of the word is clear given the full context, but it is impossible to predict given just the last sentence. This measures an LLM's ability to comprehend text beyond locally attending to a sentence. 

\textbf{Self-Supervised Approach:} We can utilize self-supervised evaluation to understand how the model's predictions change when a prior sentence or multiple sentences from a passage are altered. 
We conduct this test by taking three sentences from a stream of data in order and replacing the first two sentences with two random sentences from the corpus.
For example, if the original passage had three sentences, $\{S_3, S_2, S_1\}$, where $S_3$ is the first sentence of the input passage, then the altered passage would be $\{S_X', S_Y', S_1\}$, where $S_X', S_Y'$ are random sentences from another passage in the corpus.
A more concrete example can be found in the Appendix. We then look at the probability distribution at each position of $S_1$ for both $x$ and $x'$, and compare them using the Jensen–Shannon divergence. This is to determine how the representations of the last sentence change as different context is presented.

The Jensen-Shannon divergence ($\JSD$) is a symmetric variation of KL-divergence, defined as:
$$\JSD(P||Q)= \frac{1}{2}KL(P||M) + \frac{1}{2}KL(Q||M), \text{where } M = \frac{1}{2}(P+Q).$$

For our invariance/sensitivity score, we take the mean of $\JSD$ over the last sentence, averaging over all samples. Concretely,
$$\textsc{LRS Score} = \frac{1}{n} \sum_i^n \frac{1}{m} \sum_j^m \JSD(f(x^i_j) || f((x')^i_j)),$$
where $m$ represents the sentence length and $x^i_j$ is the $i$th sample in the set at token position $j$ in the last sentence.

\subsection{Experimental Set-up}
For this sensitivity test, we compare our method to LAMBADA using EleutherAI's Language Model Evaluation Harness \citep{eval-harness}. It is worth noting that the tests here are different. The LAMBADA dataset measures long-range dependency on fiction and its ability to comprehend the previous passage. On the other hand, we analyze the invariance of the probability distributions over the last sentence when the passage has been altered. To calculate our metric, we use the same corpus as the other tests and calculate over 1000 examples with the standard error $\expnumber{2}{-3}$ of the mean value record. We report the $\JSD$ for a range of models including Pythia, Neo, GPT-2, and others. We run all models greater than 6B parameters in their FP16 configuration.
\begin{figure}[!h]
    \centering
    \includegraphics[width=0.455\linewidth]{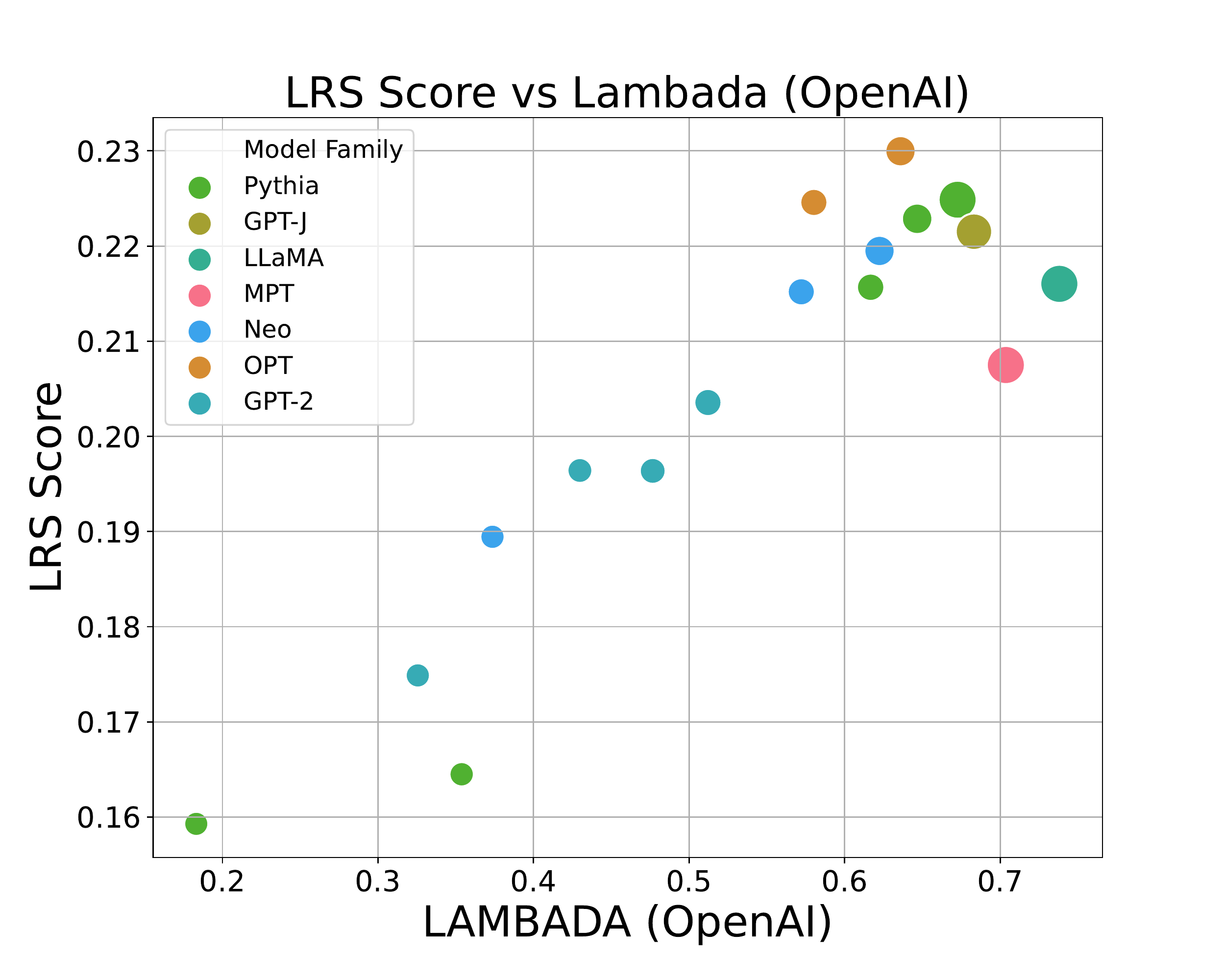}
    \includegraphics[width=0.49\linewidth]{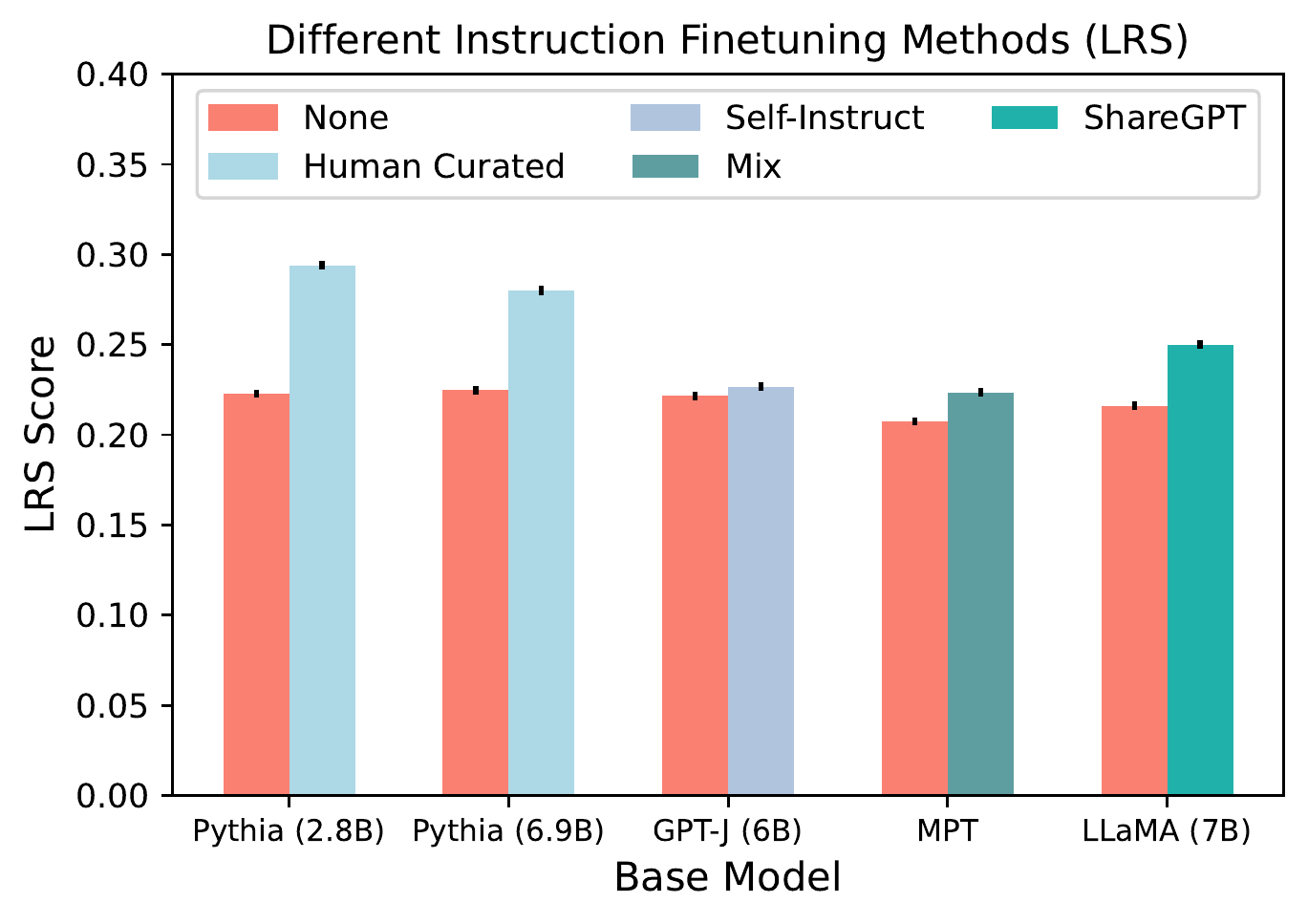}
    \caption{\textbf{Left} LRS Score vs LAMBADA (OpenAI) across various model sizes and families. \textbf{Right} LRS Score of instruction finetuned models and their pretrained counterparts.}
    \label{fig:LRD_Models_N_instruct}
    \vspace{-.25cm}
\end{figure}
\subsection{Results}
From \cref{fig:LRD_Models_N_instruct} (Left), we see that as our LRS Score increases, the model performs better on LAMBADA. Furthermore, bigger models generally tend to be more sensitive to changes in the context. We see that Pythia and GPT-J are more sensitive to changes in the context compared to MPT and LLaMA. Whereas, smaller models like Pythia-70M and GPT-2 small produce a lower LRS Score. 

\textbf{Effect of Instruction Tuning:} On average, we see that instruction-finetuned models are more sensitive to changes in context than their pretrained counterparts, suggesting that they may be sensitive to long-range changes (beyond locally attending to a sentence). Moreover, we find this gain appears independent of base model size. Both the smaller and larger Pythia base models have a similar sensitivity, and finetuning on Dolly-V2 (``human-curated'' in \cref{fig:LRD_Models_N_instruct}) leads to a similar gain in sensitivity. % This may hint at the fact that different instruction tuning datasets may impact the model differently for this metric.

\textbf{Limitations:} Although we are analyzing long-range sensitivity in token probability space, for transformers in particular, analyzing attention probabilities may be more effective. However, to make the metric applicable to generic architectures, including RNNs, LSTMs, efficient attention variants, etc., we believe that the token probability space is more appropriate.

\section{Word Order: \texttt{Word Salad Metric}} \label{sec:word_order}

\looseness -1 Close adherence to word order is a requirement for accurate factual responses beyond simple completions based on associative recall. Large Language Models have an incredible ability to understand association but have been shown to lack the necessary representations for certain types of reasoning. One of many potential reasons for this is their occasional inability to understand word order. \citet{yuksekgonul2023when} showed that multimodal models trained on image captions exhibit this behavior. People have also demonstrated that BERT can often behave like a bag-of-words classifier \citep{juneja2023linear}. 

\textbf{Self-Supervised Approach:}
\looseness -1 To evaluate a model's sensitivity to word order, we utilize sentences from a given corpus and apply a transformation where two random words are swapped in each sentence, creating modified versions denoted as $x'$. 
Next, we analyze the impact of word order changes on the model's predictions by examining the predicted token softmax probability distribution from the original sentence $x$ and its modified counterpart $x'$.
Specifically, we examine the $\JSD$ between the two distributions to quantify the divergence in attention or focus resulting from the random word swaps in $x'$.
Since there are no datasets that study word order, we compare our self-supervised approach to the LRS Score established in the previous section.

$$\textsc{Word Order Score} = \text{median}\{\JSD(f(x)_{j+1} ||f(x')_{j'+1})\text{ }\forall (x, x') \in X\},$$

where $j$ is the last token for the input sequence for $x$ and $j'$ is the last token for $x'$.
\begin{figure}[!h]
    \centering
    \includegraphics[width=0.455\linewidth]{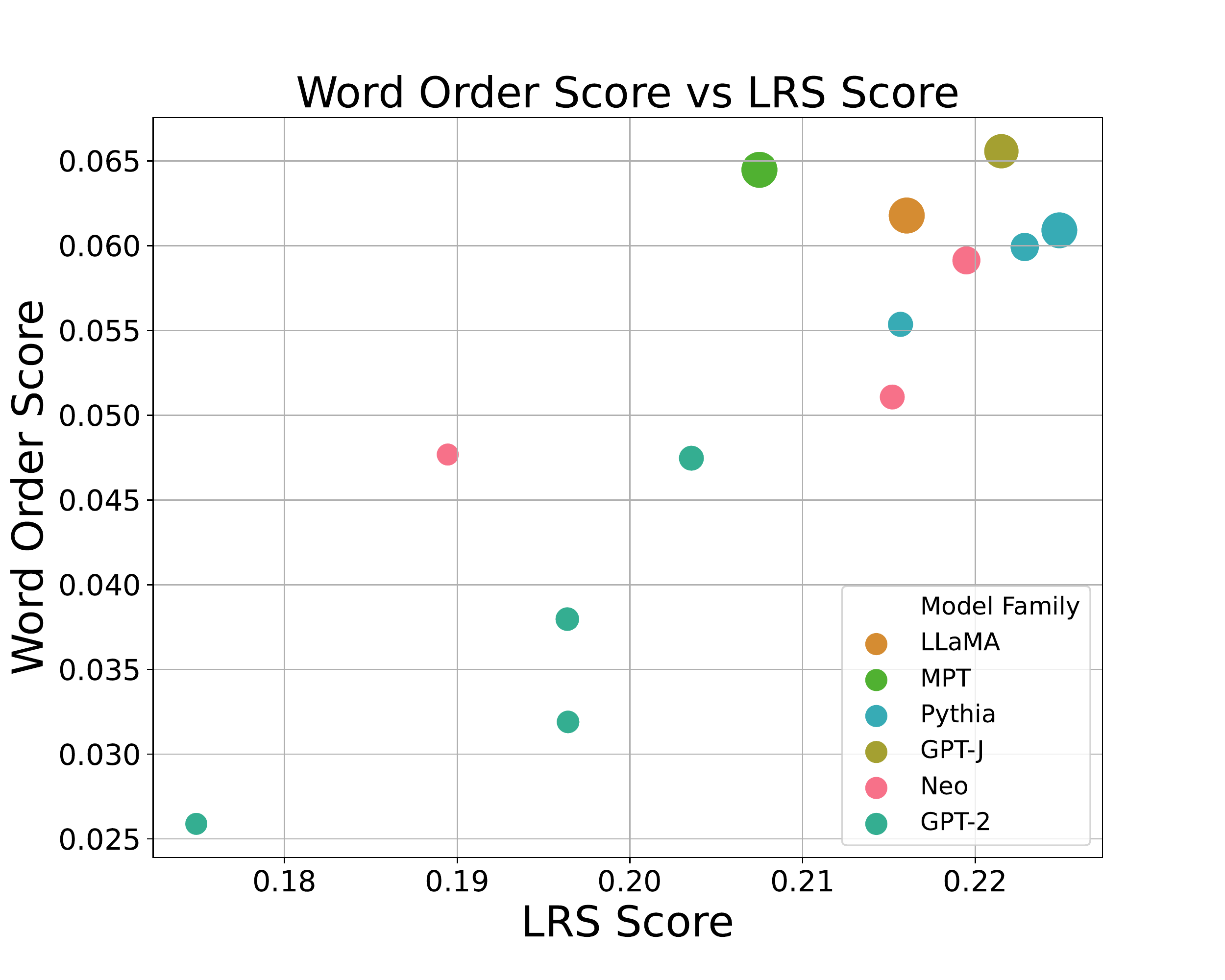}
    \includegraphics[width=0.49\linewidth]{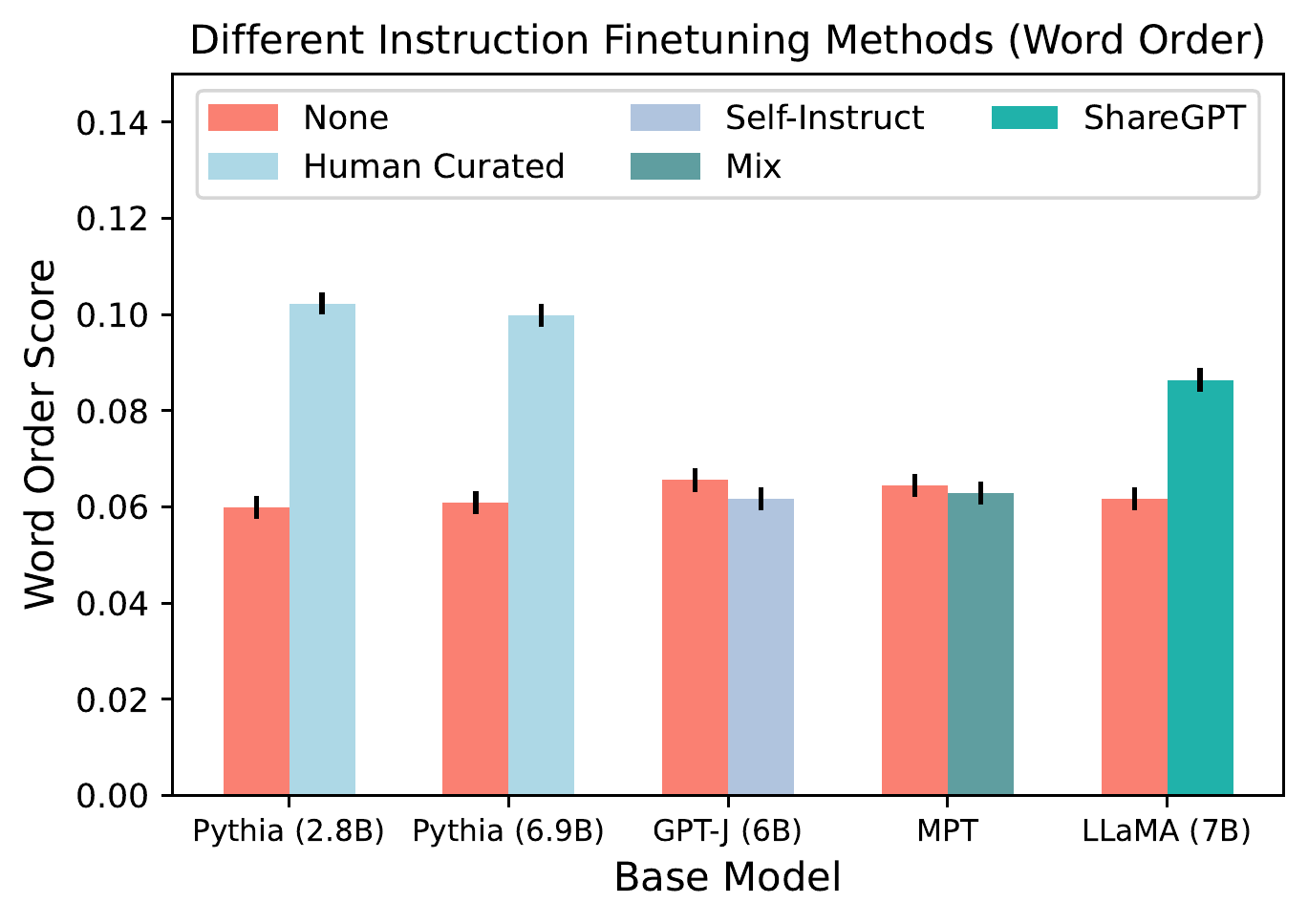}
    \caption{\textbf{(Left)} Word Order Score vs LRS Score across various model sizes and families. \textbf{(Right)} Word Order Score of instruction finetuned models and their pretrained counterparts.}
    \label{fig:wordOrder}
    \vspace{-.5cm}
\end{figure}

\subsection{Experimental Set-up}
\looseness -1 For this experiment, we take our corpus and break it down into sentences. Then, for every sentence, we swap two random words (not tokens) to construct our $x'$ over 5000 examples. 
Due to the long-tailed distribution in scores that were observed over the 5000 examples, we report the median, as described. For reference, if we had computed the mean, we would observe a standard error $\expnumber{2}{-3}$.  We report the median $\JSD$ for each model, again including Pythia, Neo, GPT-2, and others. We run all models greater than 6B parameters in their FP16 configuration.

\subsection{Results}
\looseness -1 From \cref{fig:wordOrder} (Left), we can see that there is a positive correlation between Word Order Score and LRS Score. The higher the Word Order Score, the higher the LRS Score. Nevertheless, we can see that there appears to be a plateau for Word Score.
Similar to the LRS Score, we see that larger models are more sensitive to word order, with the Mosaic MPT-7B and GPT-J model being the most sensitive to word order.

\textbf{Effect of Instruction Finetuning:} \cref{fig:wordOrder} (Right) shows that most instruction finetuning approaches make the model more sensitive to word order over the five models studied. Particularly, we see that only finetuning on the human-curated \texttt{databricks-dolly-15k} seems to make the model more sensitive irrespective of the size.

\textbf{Limitations:} For this Word Order Score, we make the assumption that the next token prediction when swapping two words randomly is a good proxy to measure a model's sensitivity to word order.

\section{Tokenization Sensitivity: \texttt{Broken Token Metric}} \label{sec:tokenization}

 Text pre-processing is rarely perfect. Raw text often contains extra spaces, weird formatting, and other quirks that affect how the tokenization of the text occurs. HELM explored some of these phenomena \citep{liang2022holistic}. Others, such as \citet{MagikarpToken}, found anomalous tokens that represent failure modes in GPT-2 and GPT-3 models, showing that our understanding of how different tokenization impacts the model behavior is still limited. 

\textbf{Self-Supervised Approach:}
\looseness -1 To quantify this phenomenon, we randomly chop strings of raw input text at regular intervals of $x$, and then we tokenize each of the chopped strings independently. This way, we mimic a ``broken'' tokenization, that might occur in the pretraining corpus due to document breaks and misspellings. A broken tokenization can also occur during model generation when incomplete user input is provided \citep{microsoft_guidance_2023}. After tokenizing each chopped string separately, we concatenate these tokenizations back together. Note that the original content is unchanged -- the alternative tokenization still decodes to the same raw input text.
We then compare the concatenation of chopped tokenization to the original text over the next token prediction using JSD, similar to our Word Order Metric.

$$\textsc{Tokenization Sensitivity Score} = \frac{1}{n}\sum\JSD(f(x)_{j+1} ||f(x')_{j'+1})$$

\subsection{Experimental Set-up}
For this experiment, we take our corpus and break it down into sentences. Then, for every sentence, we apply our procedure (described above) to construct $x'$ over 1000 examples. We report the mean $\JSD$ for each different model like Pythia, Neo, GPT-2, and others, where the standard error is about $\expnumber{5}{-3}$ for all models. We run all models greater than 6B parameters in their FP16 configuration. Here, we specifically explore a \textit{split stride} of $5$, splitting every 5th character.

\subsection{Results}
\begin{figure*}[b]
    \includegraphics[width=0.45\linewidth]{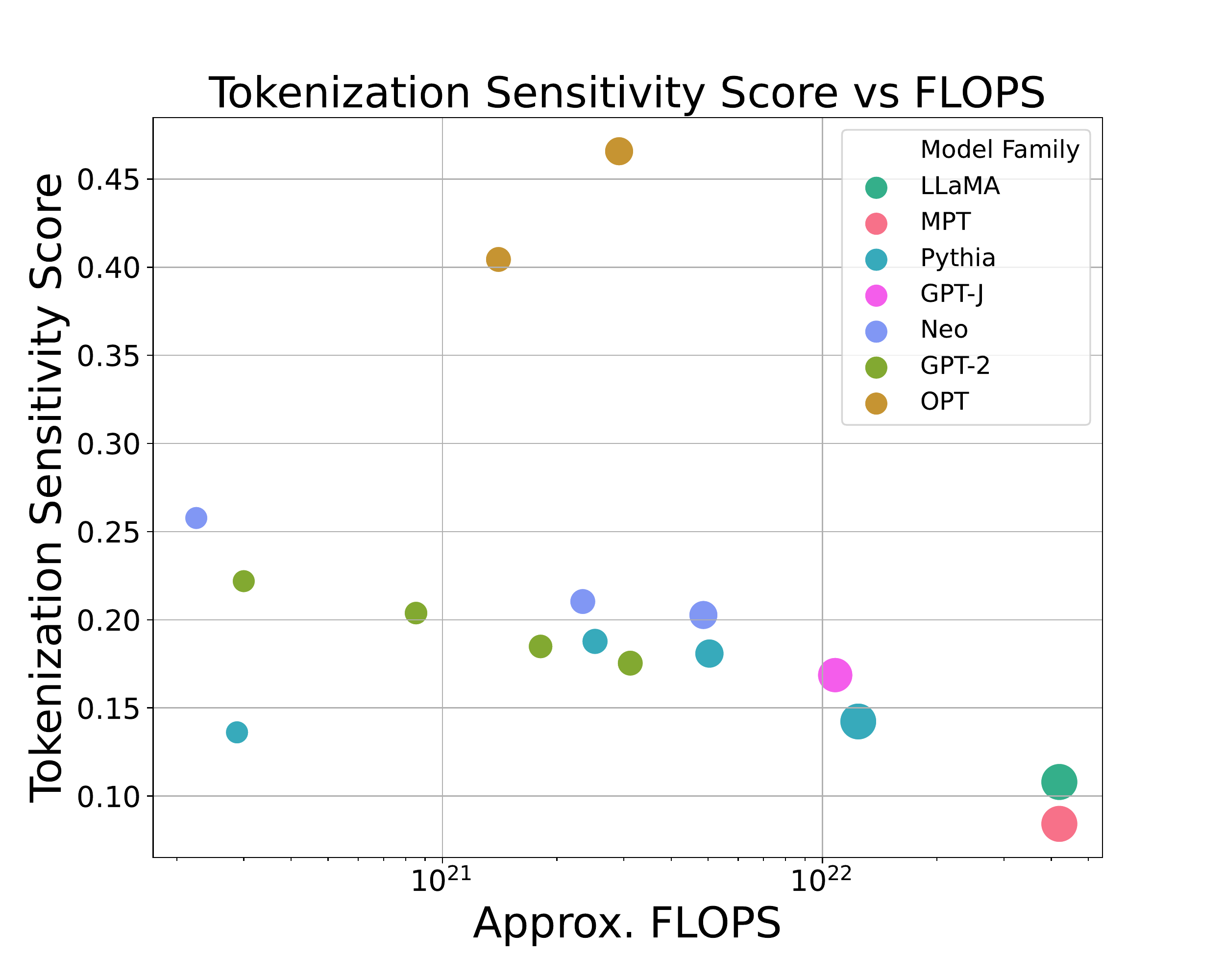}
    \includegraphics[width=0.48\linewidth]{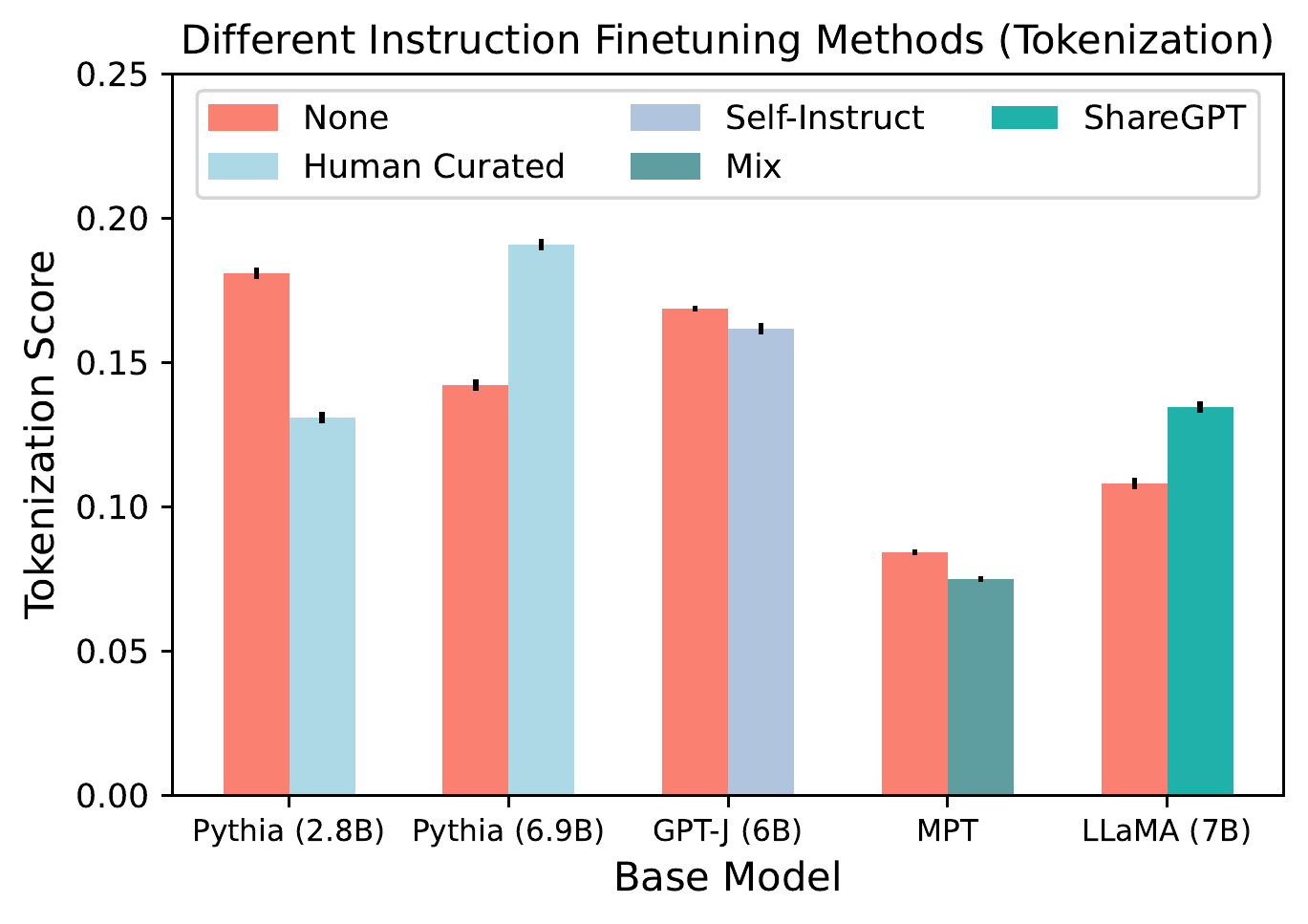}
    \caption{\textbf{(Left)} Tokenization Sensitivity Score with a split stride of five versus Approx. FLOPS -- lower is better. Note that the OPT models have seen the fewest tokens during training, c.f. \cref{fig:tokenization_tokens_seen}. \textbf{(Right)} Impact of different instruction-tuned methods.}
    \label{fig:tokenization_figs}
\end{figure*}
From \cref{fig:tokenization_figs} (Left), we see that MPT and LLaMA are the least sensitive (lower is better) to changes in token inputs. More broadly, we observe a negative trend with training FLOPs (i.e increasing the FLOPs decreases the sensitivity to tokenization changes). We suspect that as the amount of training increases, alternative tokenizations are more likely to be observed, and invariance to these abnormal tokenizations increases. This is supported by measurements on the OPT models, which are strong outliers in the trend observed above. Each of these models was trained on only 180B tokens, less than a fifth of the tokens seen by MPT and LLaMA (1 Trillion) and about half of what GPT-2, GPT-Neo, and Pythia have seen. We include \cref{fig:tokenization_tokens_seen} for a variant of \cref{fig:tokenization_figs} in terms of tokens observed during training in the appendix.  

\textbf{Effect of Instruction Finetuning:} \cref{fig:tokenization_figs} (Right) shows the impact of different instruction finetuning methods. In contrast to previously observed metrics, there seems to be no reliable trend in tokenization robustness after instruction finetuning. Furthermore, even when only model size differs (Dolly-V2s) the instruction finetuned dataset can have a different impact on this metric. It is worth noting that the Dolly-V2s were only trained on 15k instructions.

\textbf{Limitations}
We test a limited type -- character splits -- of tokenization error, particularly the same text just being processed differently by the tokenizer. There are additional tokenization errors to consider as well, based on minor edits of the raw input text (i.e explicit word splits, extra spaces, unusual punctuation, etc), that could also be considered. Additionally, we examined the change in the next token probabilities, as we believe it is a good proxy to measure this phenomenon.

\section{Discussion}
\vspace{-0.3cm}
In this paper, we introduce a new framework for \textit{Self-Supervised Evaluation} for LLMs using sensitivity (invariance) metrics. We show that sensitivity measurements like the ones explored in this paper -- knowledge via negations, toxicity, context, word order, and tokenization robustness -- can correlate with existing evaluation datasets, as we verify for the knowledge and context sensitivity metrics. We conclude that sensitivity metrics can provide meaningful insights into model behavior, which we also verify qualitatively in our study of the Cohere command model. Additionally, we see generally, except for toxicity, that larger models have better sensitivity scores compared to smaller models, mirroring other benchmarks that verify that model performance generally increases with scale. However, there are still things to consider when analyzing these models using \textit{Self-Supervised Evaluation}, which we will outline in this section. 

\looseness -1 For example, in some instances like text-ada-001 in knowledge, we see that being more sensitive is a byproduct of some other phenomena. Similarly, it may be that certain models are just insensitive in general to any transformations. This may be the case for tiny models, like the toxicity metric for GPT-2 (small) and the tokenization metric for Pythia-160M. This implies that there is a lower limit of model size where certain sensitivity metrics cease to be meaningful predictors of model qualities.

\textbf{Model Entropy.}
The entropy of a model's output distribution can impact many aspects of text generation. 
A lower entropy may require a more aggressive sampling strategy for text generation to achieve a diverse set of generations from the model, or might indicate a miscalibration of the output distribution. Similarly, the model's entropy can affect sensitivity scores. 
\begin{wrapfigure}[16]{r}{0.5\textwidth}
    \vspace{-.45cm}
    \centering
    \includegraphics[width=0.9\linewidth]{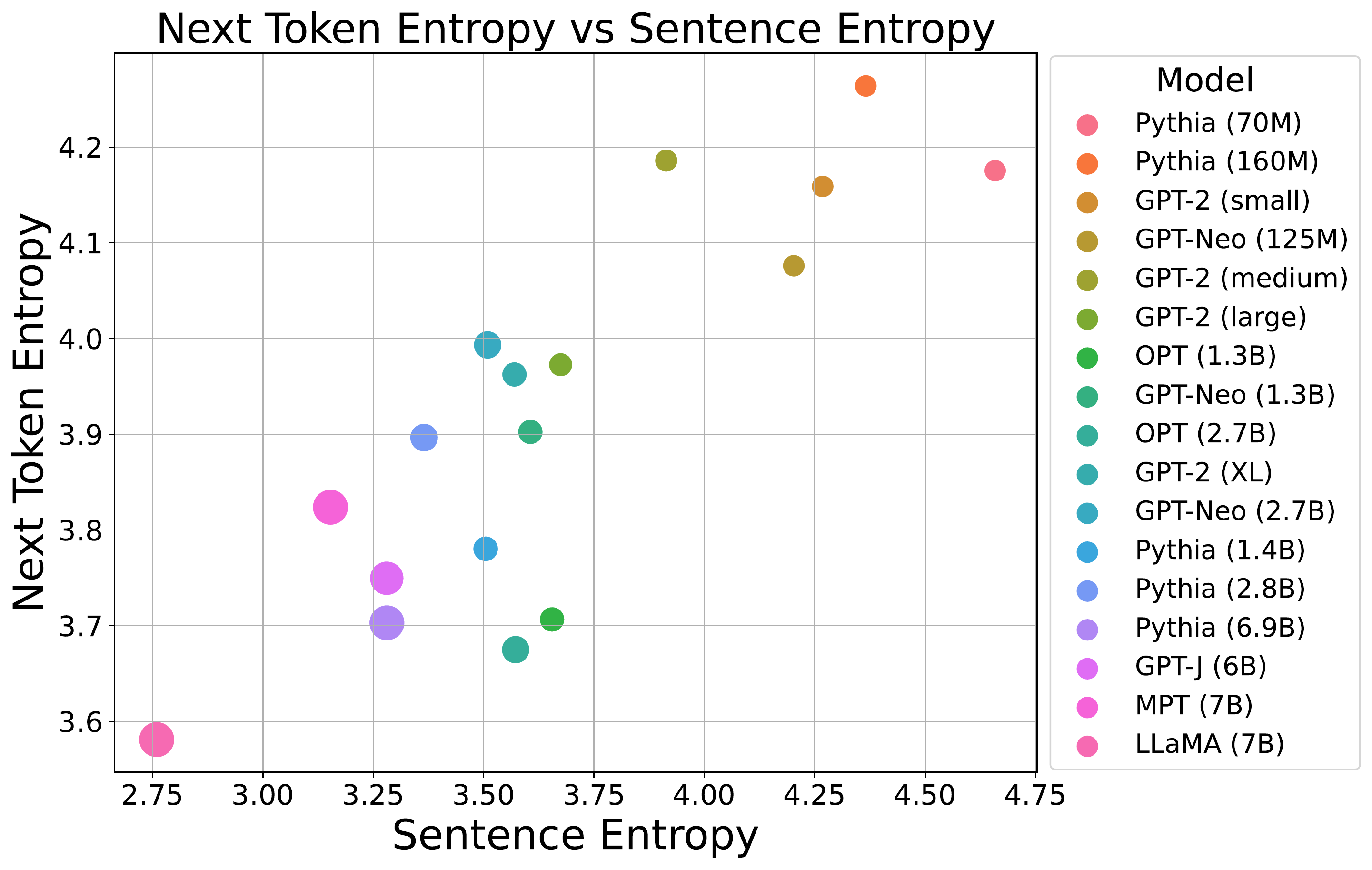}
    \caption{\looseness -1 Plot showing the next token prediction Shannon entropy (y-axis) and mean token Shannon entropy (x-axis) over sentences on Wikipedia. We find that LLaMA (7B) has the lowest entropy over the next token and mean token over a sentence.}
    \label{fig:entropy}
\end{wrapfigure}
If the entropy of the model is low, then the sensitivity may naturally be lower as well. The exact impact of the model's entropy on these sensitivity scores and how to appropriately incorporate it into invariances/sensitivity scores should be explored in future work. 
\cref{fig:entropy} shows the Shannon Entropy of the Next Token Prediction and Sentence Entropy (the mean token entropy over a sentence of the model). We use the Wikipedia (our corpus) sentences to calculate the Shannon Entropy, defined as $H(x) = -\sum p(x)\log(p(x))$. From \cref{fig:entropy}, we see that LLaMA has the lowest entropy on both the next token and mean token over a sentence, with large models having a lower entropy than smaller models on average. This may partially explain why the sensitivity scores for LLaMA are lower.
\footnote{Vocabulary size does play an additional role in the entropy of a model. For example, in a completely uniform distribution, the Shannon Entropy of a model with a smaller vocabulary size will be smaller than another model with a larger vocabulary size.}

\textbf{Memorization.}
Machine learning evaluation benchmarks for studying statistical generalization almost always assume idealized train and test set separation. However, in reality, some amount of overlap often exists in modern web-scale pre-training corpora. As a result, there have been various efforts to measure and address the impact of these overlaps on the training and evaluation of large models \citep{brown2020language,eval-harness}. Investigating the same relationship, purely from a training support perspective, \citet{kandpal2022large} showed that a language model’s ability to answer a fact-based question relates to how many documents associated with that question were seen during pre-training. In a different but fundamentally related line of work, \citet{carlini2022quantifying} demonstrated that LLMs regurgitate training data in specific scenarios, often based on repetition rates in training corpora. Further, their own prior work \citep{carlini2020extracting} quantifies the underlying relationship between train and test data in yet another way by showing that simple loss-based membership inference methods are capable of discriminating whether a test query was present in the training dataset. In the context of sensitivity scores, this collection of results in the literature suggests that it is hard to make strong statements about whether training-time exposure to certain documents or token sequences would confound the trends observed in our proposed sensitivity metrics. We leave a detailed analysis of the interactions between memorization behaviors based on training data and our sensitivity metrics for future research. We suspect that developing a more complete understanding of these interactions is an important step towards more informative and robust sensitivity metrics. 

An advantage of self-supervised sensitivity scores is that we can circumvent the potential effects of memorization by evaluating sensitivities on novel text, i.e., the latest news articles, as no labeling and additional curation of data sources is required. With this strategy, the possibility of memorization can be eliminated. 

\section{Conclusion}
\begin{figure}[t]
    \centering
    \includegraphics[width=0.45\linewidth]{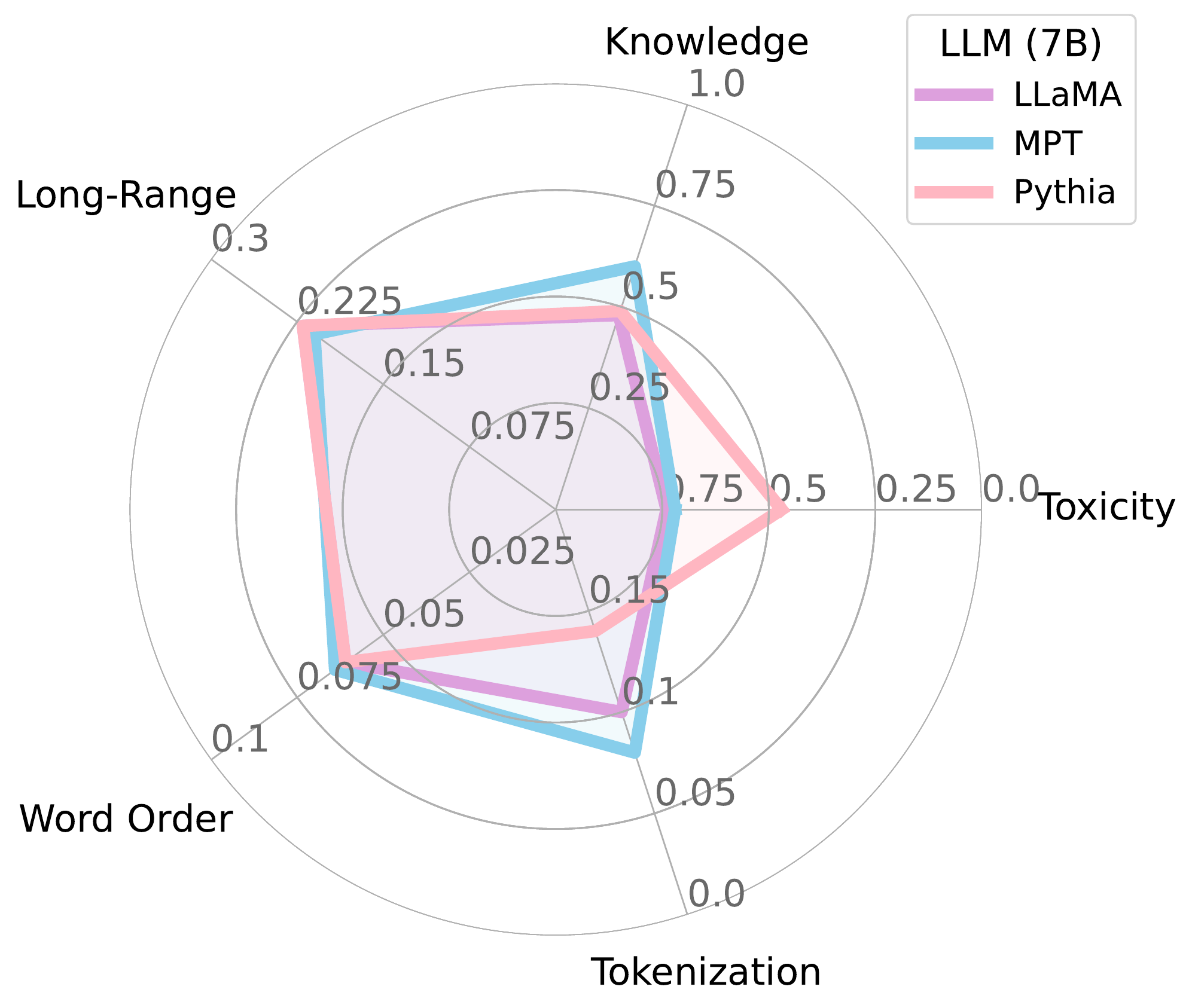}
    \includegraphics[width=0.45\linewidth]{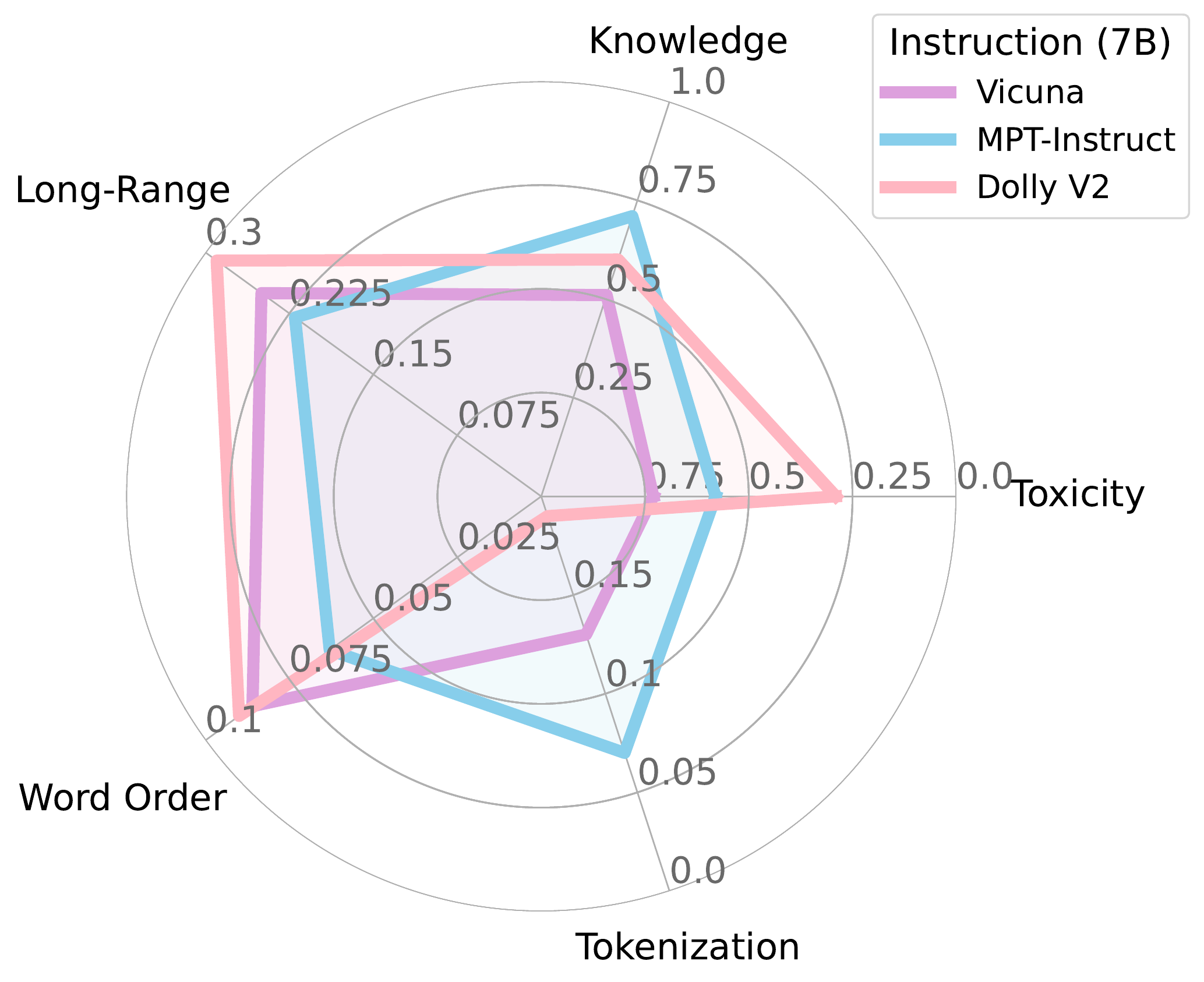}
    \caption{\textbf{(Left)} shows LLaMA, MPT, and Pythia sensitivity scores across the five metrics studied in this paper. \textbf{(Right)} shows the instruction-tuned counterparts of these models across the five metrics. The more area that is covered, the better the model according to our SSE scheme. All models 7B were run in FP16 configurations.}
    \vspace{-.25cm}
    \label{fig:spiderplots}
\end{figure}
\looseness -1 We introduce a procedure for self-supervised evaluation by analyzing invariances for Large Language Models. The key advantage of self-supervised evaluation is that it removes the need to laboriously label new data, leading to more efficient forms of evaluation in real deployment settings. We showcase several case studies, where we empirically validate this approach to be reliably tracking existing supervised metrics. Additionally, there are a number of future questions to consider when measuring a model's sensitivity that we did not fully explore yet -- like entropy and memorization.
Nevertheless, these self-supervised evaluation approaches have the potential to measure properties beyond what is currently capable of the traditional dataset approach -- like sensitivity to word order. We hope that this is \textit{only a starting point} for self-supervised metrics in the future that can lead to a deeper understanding of how LLMs behave and complement classical supervised benchmarks.

\section{Acknowledgements}
This work was made possible by the ONR MURI program, the Office of Naval Research (N000142112557), and the AFOSR MURI program. Commercial support was provided by Capital One Bank, the Amazon Research Award program, and Open Philanthropy. Further support was provided by the National Science Foundation (IIS-2212182), and by the NSF TRAILS Institute (2229885).

\bibliography{references}

\begin{thebibliography}{41}
\providecommand{\natexlab}[1]{#1}
\providecommand{\url}[1]{\texttt{#1}}
\expandafter\ifx\csname urlstyle\endcsname\relax
  \providecommand{\doi}[1]{doi: #1}\else
  \providecommand{\doi}{doi: \begingroup \urlstyle{rm}\Url}\fi

\bibitem[Anthropic(2023{\natexlab{a}})]{Anthropic_2023}
Anthropic.
\newblock Introducing 100k context windows, May 2023{\natexlab{a}}.
\newblock URL \url{https://www.anthropic.com/index/100k-context-windows}.

\bibitem[Anthropic(2023{\natexlab{b}})]{Anthropic_Claude_2023}
Anthropic.
\newblock Introducing claude, March 2023{\natexlab{b}}.
\newblock URL \url{https://www.anthropic.com/index/introducing-claude}.

\bibitem[Bai et~al.(2022)Bai, Jones, Ndousse, Askell, Chen, DasSarma, Drain,
  Fort, Ganguli, Henighan, et~al.]{bai2022training}
Yuntao Bai, Andy Jones, Kamal Ndousse, Amanda Askell, Anna Chen, Nova DasSarma,
  Dawn Drain, Stanislav Fort, Deep Ganguli, Tom Henighan, et~al.
\newblock Training a helpful and harmless assistant with reinforcement learning
  from human feedback.
\newblock \emph{arXiv preprint arXiv:2204.05862}, 2022.

\bibitem[Biderman et~al.(2023)Biderman, Schoelkopf, Anthony, Bradley, O'Brien,
  Hallahan, Khan, Purohit, Prashanth, Raff, et~al.]{biderman2023pythia}
Stella Biderman, Hailey Schoelkopf, Quentin Anthony, Herbie Bradley, Kyle
  O'Brien, Eric Hallahan, Mohammad~Aflah Khan, Shivanshu Purohit, USVSN~Sai
  Prashanth, Edward Raff, et~al.
\newblock Pythia: A suite for analyzing large language models across training
  and scaling.
\newblock \emph{arXiv preprint arXiv:2304.01373}, 2023.

\bibitem[Birhane et~al.(2022)Birhane, Kalluri, Card, Agnew, Dotan, and
  Bao]{birhane2022values}
Abeba Birhane, Pratyusha Kalluri, Dallas Card, William Agnew, Ravit Dotan, and
  Michelle Bao.
\newblock The values encoded in machine learning research.
\newblock In \emph{2022 ACM Conference on Fairness, Accountability, and
  Transparency}, pages 173--184, 2022.

\bibitem[Bowman and Dahl(2021)]{bowman2021will}
Samuel~R Bowman and George~E Dahl.
\newblock What will it take to fix benchmarking in natural language
  understanding?
\newblock \emph{arXiv preprint arXiv:2104.02145}, 2021.

\bibitem[Brown et~al.(2020)Brown, Mann, Ryder, Subbiah, Kaplan, Dhariwal,
  Neelakantan, Shyam, Sastry, Askell, et~al.]{brown2020language}
Tom Brown, Benjamin Mann, Nick Ryder, Melanie Subbiah, Jared~D Kaplan, Prafulla
  Dhariwal, Arvind Neelakantan, Pranav Shyam, Girish Sastry, Amanda Askell,
  et~al.
\newblock Language models are few-shot learners.
\newblock \emph{Advances in neural information processing systems},
  33:\penalty0 1877--1901, 2020.

\bibitem[Carlini et~al.(2020)Carlini, Tramer, Wallace, Jagielski, Herbert-Voss,
  Lee, Roberts, Brown, Song, Erlingsson, et~al.]{carlini2020extracting}
N~Carlini, F~Tramer, E~Wallace, M~Jagielski, A~Herbert-Voss, K~Lee, A~Roberts,
  T~Brown, D~Song, {\'U}~Erlingsson, et~al.
\newblock Extracting training data from large language models. arxiv.
\newblock \emph{Preprint posted online December}, 14, 2020.

\bibitem[Carlini et~al.(2022)Carlini, Ippolito, Jagielski, Lee, Tramer, and
  Zhang]{carlini2022quantifying}
Nicholas Carlini, Daphne Ippolito, Matthew Jagielski, Katherine Lee, Florian
  Tramer, and Chiyuan Zhang.
\newblock Quantifying memorization across neural language models.
\newblock \emph{arXiv preprint arXiv:2202.07646}, 2022.

\bibitem[Dao et~al.(2022)Dao, Fu, Ermon, Rudra, and Ré]{dao2022flashattention}
Tri Dao, Daniel~Y. Fu, Stefano Ermon, Atri Rudra, and Christopher Ré.
\newblock Flashattention: Fast and memory-efficient exact attention with
  io-awareness, 2022.

\bibitem[Dhamala et~al.(2021)Dhamala, Sun, Kumar, Krishna, Pruksachatkun,
  Chang, and Gupta]{BOLD}
Jwala Dhamala, Tony Sun, Varun Kumar, Satyapriya Krishna, Yada Pruksachatkun,
  Kai-Wei Chang, and Rahul Gupta.
\newblock Bold: Dataset and metrics for measuring biases in open-ended language
  generation.
\newblock In \emph{Proceedings of the 2021 ACM Conference on Fairness,
  Accountability, and Transparency}, FAccT '21, page 862–872, New York, NY,
  USA, 2021. Association for Computing Machinery.
\newblock ISBN 9781450383097.
\newblock \doi{10.1145/3442188.3445924}.
\newblock URL \url{https://doi.org/10.1145/3442188.3445924}.

\bibitem[Dhole et~al.(2021)Dhole, Gangal, Gehrmann, Gupta, Li, Mahamood,
  Mahendiran, Mille, Shrivastava, Tan, et~al.]{dhole2021nl}
Kaustubh~D Dhole, Varun Gangal, Sebastian Gehrmann, Aadesh Gupta, Zhenhao Li,
  Saad Mahamood, Abinaya Mahendiran, Simon Mille, Ashish Shrivastava, Samson
  Tan, et~al.
\newblock Nl-augmenter: A framework for task-sensitive natural language
  augmentation.
\newblock \emph{arXiv preprint arXiv:2112.02721}, 2021.

\bibitem[Ethayarajh and Jurafsky(2020)]{ethayarajh2020utility}
Kawin Ethayarajh and Dan Jurafsky.
\newblock Utility is in the eye of the user: A critique of {NLP} leaderboards.
\newblock \emph{arXiv preprint arXiv:2009.13888}, 2020.

\bibitem[Ettinger(2020)]{ettinger2020bert}
Allyson Ettinger.
\newblock What bert is not: Lessons from a new suite of psycholinguistic
  diagnostics for language models.
\newblock \emph{Transactions of the Association for Computational Linguistics},
  8:\penalty0 34--48, 2020.

\bibitem[Fortuna et~al.(2020)Fortuna, Soler, and
  Wanner]{fortuna-etal-2020-toxic}
Paula Fortuna, Juan Soler, and Leo Wanner.
\newblock Toxic, hateful, offensive or abusive? what are we really classifying?
  an empirical analysis of hate speech datasets.
\newblock In \emph{Proceedings of the Twelfth Language Resources and Evaluation
  Conference}, pages 6786--6794, Marseille, France, May 2020. European Language
  Resources Association.
\newblock ISBN 979-10-95546-34-4.
\newblock URL \url{https://aclanthology.org/2020.lrec-1.838}.

\bibitem[Gao et~al.(2021)Gao, Tow, Biderman, Black, DiPofi, Foster, Golding,
  Hsu, McDonell, Muennighoff, Phang, Reynolds, Tang, Thite, Wang, Wang, and
  Zou]{eval-harness}
Leo Gao, Jonathan Tow, Stella Biderman, Sid Black, Anthony DiPofi, Charles
  Foster, Laurence Golding, Jeffrey Hsu, Kyle McDonell, Niklas Muennighoff,
  Jason Phang, Laria Reynolds, Eric Tang, Anish Thite, Ben Wang, Kevin Wang,
  and Andy Zou.
\newblock A framework for few-shot language model evaluation, September 2021.
\newblock URL \url{https://doi.org/10.5281/zenodo.5371628}.

\bibitem[Gehman et~al.(2020)Gehman, Gururangan, Sap, Choi, and
  Smith]{gehman_realtoxicityprompts_2020}
Samuel Gehman, Suchin Gururangan, Maarten Sap, Yejin Choi, and Noah~A. Smith.
\newblock {RealToxicityPrompts}: {Evaluating} {Neural} {Toxic} {Degeneration}
  in {Language} {Models}, September 2020.
\newblock URL \url{http://arxiv.org/abs/2009.11462}.
\newblock arXiv:2009.11462 [cs].

\bibitem[Hendrycks et~al.(2021)Hendrycks, Burns, Basart, Zou, Mazeika, Song,
  and Steinhardt]{hendrycks2021measuring}
Dan Hendrycks, Collin Burns, Steven Basart, Andy Zou, Mantas Mazeika, Dawn
  Song, and Jacob Steinhardt.
\newblock Measuring massive multitask language understanding, 2021.

\bibitem[Joshi et~al.(2017)Joshi, Choi, Weld, and
  Zettlemoyer]{joshi2017triviaqa}
Mandar Joshi, Eunsol Choi, Daniel~S Weld, and Luke Zettlemoyer.
\newblock Triviaqa: A large scale distantly supervised challenge dataset for
  reading comprehension.
\newblock In \emph{Proceedings of the 55th Annual Meeting of the Association
  for Computational Linguistics (Volume 1: Long Papers)}, pages 1601--1611,
  2017.

\bibitem[Juneja et~al.(2023)Juneja, Bansal, Cho, Sedoc, and
  Saphra]{juneja2023linear}
Jeevesh Juneja, Rachit Bansal, Kyunghyun Cho, Jo{\~a}o Sedoc, and Naomi Saphra.
\newblock Linear connectivity reveals generalization strategies.
\newblock In \emph{The Eleventh International Conference on Learning
  Representations}, 2023.
\newblock URL \url{https://openreview.net/forum?id=hY6M0JHl3uL}.

\bibitem[Kandpal et~al.(2022)Kandpal, Deng, Roberts, Wallace, and
  Raffel]{kandpal2022large}
Nikhil Kandpal, Haikang Deng, Adam Roberts, Eric Wallace, and Colin Raffel.
\newblock Large language models struggle to learn long-tail knowledge.
\newblock \emph{arXiv preprint arXiv:2211.08411}, 2022.

\bibitem[Kiela et~al.(2021)Kiela, Bartolo, Nie, Kaushik, Geiger, Wu, Vidgen,
  Prasad, Singh, Ringshia, et~al.]{kiela2021dynabench}
Douwe Kiela, Max Bartolo, Yixin Nie, Divyansh Kaushik, Atticus Geiger,
  Zhengxuan Wu, Bertie Vidgen, Grusha Prasad, Amanpreet Singh, Pratik Ringshia,
  et~al.
\newblock Dynabench: Rethinking benchmarking in nlp.
\newblock \emph{arXiv preprint arXiv:2104.14337}, 2021.

\bibitem[Liang et~al.(2022)Liang, Bommasani, Lee, Tsipras, Soylu, Yasunaga,
  Zhang, Narayanan, Wu, Kumar, et~al.]{liang2022holistic}
Percy Liang, Rishi Bommasani, Tony Lee, Dimitris Tsipras, Dilara Soylu,
  Michihiro Yasunaga, Yian Zhang, Deepak Narayanan, Yuhuai Wu, Ananya Kumar,
  et~al.
\newblock Holistic evaluation of language models.
\newblock \emph{arXiv preprint arXiv:2211.09110}, 2022.

\bibitem[Microsoft(2023)]{microsoft_guidance_2023}
Microsoft.
\newblock Guidance.
\newblock Microsoft, June 2023.
\newblock URL \url{https://github.com/microsoft/guidance}.

\bibitem[Mille et~al.(2021)Mille, Dhole, Mahamood, Perez-Beltrachini, Gangal,
  Kale, van Miltenburg, and Gehrmann]{Mille2021AutomaticCO}
Simon Mille, Kaustubh~D. Dhole, Saad Mahamood, Laura Perez-Beltrachini, Varun
  Gangal, Mihir Kale, Emiel van Miltenburg, and Sebastian Gehrmann.
\newblock Automatic construction of evaluation suites for natural language
  generation datasets.
\newblock \emph{ArXiv}, abs/2106.09069, 2021.

\bibitem[O'Connor and Andreas(2021)]{oconnor2021context}
Joe O'Connor and Jacob Andreas.
\newblock What context features can transformer language models use?
\newblock \emph{arXiv preprint arXiv:2106.08367}, 2021.

\bibitem[OpenAI(2023)]{openai2023gpt4}
OpenAI.
\newblock Gpt-4 technical report.
\newblock \emph{ArXiv}, abs/2303.08774, 2023.

\bibitem[Ouyang et~al.(2022)Ouyang, Wu, Jiang, Almeida, Wainwright, Mishkin,
  Zhang, Agarwal, Slama, Ray, et~al.]{ouyang2022training}
Long Ouyang, Jeffrey Wu, Xu~Jiang, Diogo Almeida, Carroll Wainwright, Pamela
  Mishkin, Chong Zhang, Sandhini Agarwal, Katarina Slama, Alex Ray, et~al.
\newblock Training language models to follow instructions with human feedback.
\newblock \emph{Advances in Neural Information Processing Systems},
  35:\penalty0 27730--27744, 2022.

\bibitem[Paperno et~al.(2016)Paperno, Kruszewski, Lazaridou, Pham, Bernardi,
  Pezzelle, Baroni, Boleda, and Fern{\'a}ndez]{paperno2016lambada}
Denis Paperno, Germ{\'a}n Kruszewski, Angeliki Lazaridou, Ngoc-Quan Pham,
  Raffaella Bernardi, Sandro Pezzelle, Marco Baroni, Gemma Boleda, and Raquel
  Fern{\'a}ndez.
\newblock The lambada dataset: Word prediction requiring a broad discourse
  context.
\newblock In \emph{Proceedings of the 54th Annual Meeting of the Association
  for Computational Linguistics (Volume 1: Long Papers)}, pages 1525--1534,
  2016.

\bibitem[Pozzobon et~al.(2023)Pozzobon, Ermis, Lewis, and
  Hooker]{pozzobon2023challenges}
Luiza Pozzobon, Beyza Ermis, Patrick Lewis, and Sara Hooker.
\newblock On the challenges of using black-box apis for toxicity evaluation in
  research.
\newblock \emph{arXiv preprint arXiv:2304.12397}, 2023.

\bibitem[Radford et~al.(2019)Radford, Wu, Child, Luan, Amodei, Sutskever,
  et~al.]{radford2019language}
Alec Radford, Jeffrey Wu, Rewon Child, David Luan, Dario Amodei, Ilya
  Sutskever, et~al.
\newblock Language models are unsupervised multitask learners.
\newblock \emph{OpenAI blog}, 1\penalty0 (8):\penalty0 9, 2019.

\bibitem[Ribeiro et~al.(2020)Ribeiro, Wu, Guestrin, and
  Singh]{ribeiro-etal-2020-beyond}
Marco~Tulio Ribeiro, Tongshuang Wu, Carlos Guestrin, and Sameer Singh.
\newblock Beyond accuracy: Behavioral testing of {NLP} models with
  {C}heck{L}ist.
\newblock In \emph{Proceedings of the 58th Annual Meeting of the Association
  for Computational Linguistics}, pages 4902--4912, Online, July 2020.
  Association for Computational Linguistics.
\newblock \doi{10.18653/v1/2020.acl-main.442}.
\newblock URL \url{https://aclanthology.org/2020.acl-main.442}.

\bibitem[Ross et~al.(2021)Ross, Wu, Peng, Peters, and
  Gardner]{Ross2021TailorGA}
Alexis Ross, Tongshuang~Sherry Wu, Hao Peng, Matthew~E. Peters, and Matt
  Gardner.
\newblock Tailor: Generating and perturbing text with semantic controls.
\newblock In \emph{Annual Meeting of the Association for Computational
  Linguistics}, 2021.

\bibitem[Rumbelow and Mwatkins(2023)]{MagikarpToken}
Jessica Rumbelow and Mwatkins.
\newblock Solidgoldmagikarp (plus, prompt generation), 2023.
\newblock URL
  \url{https://www.lesswrong.com/posts/aPeJE8bSo6rAFoLqg/solidgoldmagikarp-plus-prompt-generation}.

\bibitem[Srivastava et~al.(2022)Srivastava, Rastogi, Rao, Shoeb, Abid, Fisch,
  Brown, Santoro, Gupta, Garriga-Alonso, et~al.]{srivastava2022beyond}
Aarohi Srivastava, Abhinav Rastogi, Abhishek Rao, Abu Awal~Md Shoeb, Abubakar
  Abid, Adam Fisch, Adam~R Brown, Adam Santoro, Aditya Gupta, Adri{\`a}
  Garriga-Alonso, et~al.
\newblock Beyond the imitation game: Quantifying and extrapolating the
  capabilities of language models.
\newblock \emph{arXiv preprint arXiv:2206.04615}, 2022.

\bibitem[Sun et~al.(2022)Sun, Xu, Deng, Cheng, Zheng, Zhou, Peng, Zhu, and
  Huang]{sun-etal-2022-safety}
Hao Sun, Guangxuan Xu, Jiawen Deng, Jiale Cheng, Chujie Zheng, Hao Zhou, Nanyun
  Peng, Xiaoyan Zhu, and Minlie Huang.
\newblock On the safety of conversational models: Taxonomy, dataset, and
  benchmark.
\newblock In \emph{Findings of the Association for Computational Linguistics:
  ACL 2022}, pages 3906--3923, Dublin, Ireland, May 2022. Association for
  Computational Linguistics.
\newblock \doi{10.18653/v1/2022.findings-acl.308}.
\newblock URL \url{https://aclanthology.org/2022.findings-acl.308}.

\bibitem[Tay et~al.(2020)Tay, Dehghani, Abnar, Shen, Bahri, Pham, Rao, Yang,
  Ruder, and Metzler]{taylong}
Yi~Tay, Mostafa Dehghani, Samira Abnar, Yikang Shen, Dara Bahri, Philip Pham,
  Jinfeng Rao, Liu Yang, Sebastian Ruder, and Donald Metzler.
\newblock Long range arena: A benchmark for efficient transformers.
\newblock In \emph{International Conference on Learning Representations}, 2020.

\bibitem[Thrush et~al.(2022)Thrush, Jiang, Bartolo, Singh, Williams, Kiela, and
  Ross]{thrush2022winoground}
Tristan Thrush, Ryan Jiang, Max Bartolo, Amanpreet Singh, Adina Williams, Douwe
  Kiela, and Candace Ross.
\newblock Winoground: Probing vision and language models for visio-linguistic
  compositionality, 2022.

\bibitem[Wu et~al.(2021)Wu, Ribeiro, Heer, and Weld]{Wu2021PolyjuiceGC}
Tongshuang~Sherry Wu, Marco~Tulio Ribeiro, Jeffrey Heer, and Daniel~S. Weld.
\newblock Polyjuice: Generating counterfactuals for explaining, evaluating, and
  improving models.
\newblock In \emph{Annual Meeting of the Association for Computational
  Linguistics}, 2021.

\bibitem[Yang et~al.(2022)Yang, Haque, Song, Yang, and Liu]{Yang2022TestAugAF}
Guanqun Yang, Mirazul Haque, Qiaochu Song, Wei Yang, and Xueqing Liu.
\newblock Testaug: A framework for augmenting capability-based nlp tests.
\newblock In \emph{International Conference on Computational Linguistics},
  2022.

\bibitem[Yuksekgonul et~al.(2023)Yuksekgonul, Bianchi, Kalluri, Jurafsky, and
  Zou]{yuksekgonul2023when}
Mert Yuksekgonul, Federico Bianchi, Pratyusha Kalluri, Dan Jurafsky, and James
  Zou.
\newblock When and why vision-language models behave like bags-of-words, and
  what to do about it?
\newblock In \emph{The Eleventh International Conference on Learning
  Representations}, 2023.
\newblock URL \url{https://openreview.net/forum?id=KRLUvxh8uaX}.

\end{thebibliography}
\bibliographystyle{plainnat}
\clearpage
\appendix
\section{Appendix}
\subsection{Knowledge Probing via Negations}
\begin{figure}
    \centering
    \includegraphics[width=0.45\linewidth]{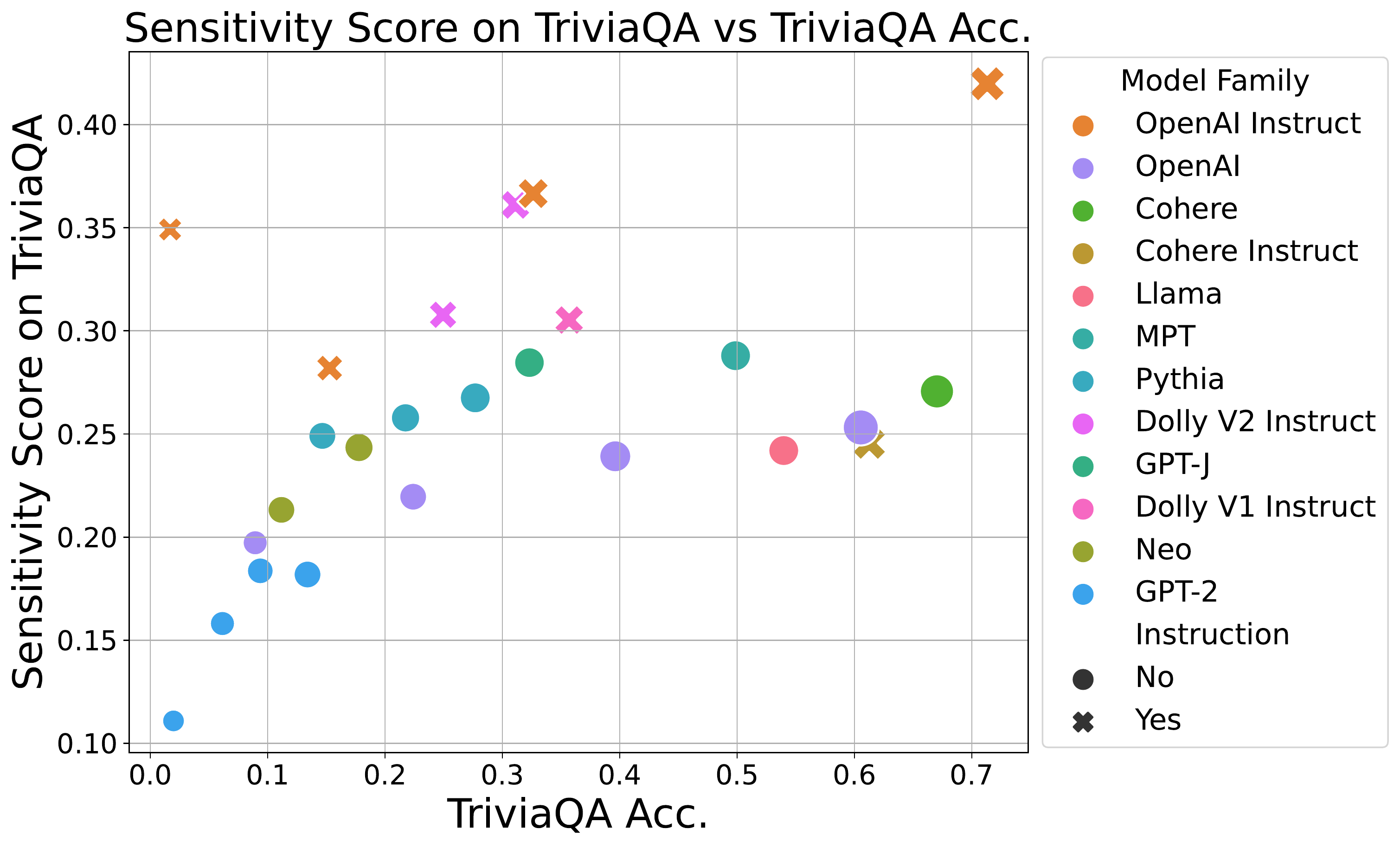}
    \includegraphics[width=0.45\linewidth]{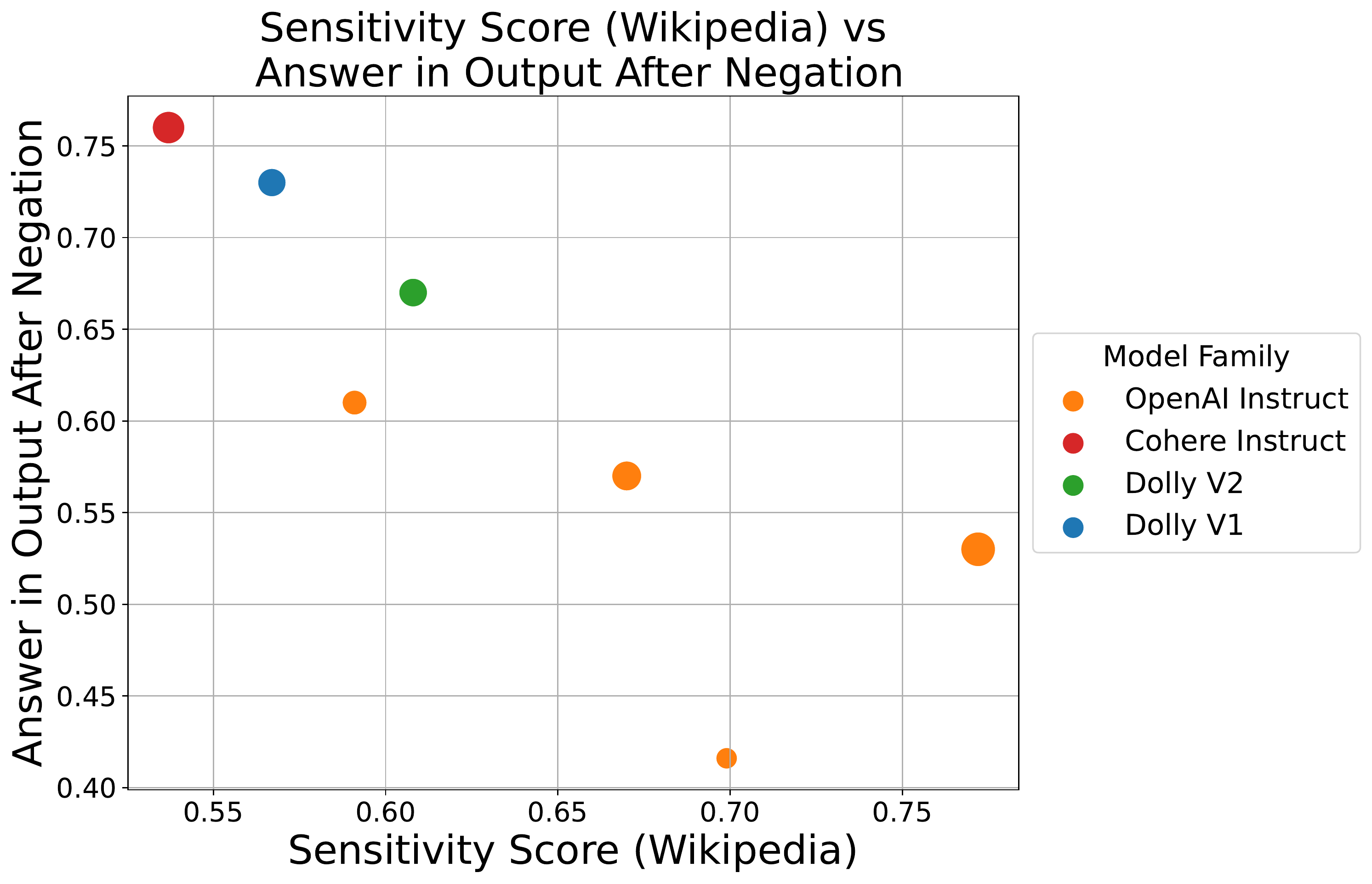}
    \caption{\textbf{(Left)} The change in perplexity in the question-answer pair when a negation is applied to the question versus TriviaQA Acc. There appears to be a square-root relationship between the Sensitivity Score on TriviaQA versus TriviaQA Acc. \textbf{(Right)} The percentage of times when the correct answer was contained in the solution even when applying the negation versus Sensitivity Score (Wikipedia) for a few instruction models. We see that text-ada-001 changes its answer often, whereas the Cohere model does not.}
    \label{fig:neg_controlled_triviaQA}
\end{figure}

\paragraph{Example:} \cref{fig:negatiom_example} shows an example of the original $x$ and the transformed $x'$ for the \emph{Knowledge Probing via Negations} experiments.
\begin{figure}[h]
\begin{center}
\noindent\fbox{%
    \parbox{0.8\textwidth}{%
\textbf{Original ($x$):} April is the fourth month of the year in the Julian and Gregorian calendars and comes between March and May. \\\\
\textbf{Perturbed ($x'$):} April is \underline{not} the fourth month of the year in the Julian and Gregorian calendars and comes between March and May.} }
\end{center}
    \caption{Knowledge probing via negations example over topic sentences in \texttt{wikipedia}. \textbf{(Top)} is the original, $x$, from \texttt{wikipedia}. \textbf{(Bottom)} is the transformed, $x'$, where we add a ``not'' according to the rules described in the main paper.}
    \label{fig:negatiom_example}
\end{figure}
\paragraph{Adding Negations in TriviaQA}
To understand whether adding negations and measuring the change in log perplexity is a reasonable assessment of probing the knowledge in an LLM, we added negations to questions following the same rule described in the main paper. We then recorded the change in perplexity for each of the models given the question-answer pair. 
This was to understand how different models may understand negations. \cref{fig:neg_controlled_triviaQA} (Left) shows that adding a negation in the question and observing the change in perplexity can give us an indication of performance on TriviaQA. 

\paragraph{TriviaQA Accuracy}
We calculate the accuracy for TriviaQA for the \texttt{unfiltered-web-dev} split by simply counting a correct answer from the model if one of the given answers was contained in the output string. Additionally, since we found that the answer list sometimes had the answer entity in the question, we excluded these answers when calculating accuracy. We use the template ``Question: \texttt{[input question]}\textbackslash nAnswer:''.

\looseness -1  \textbf{Models From Huggingface:} \texttt{gpt2, gpt2-large, gpt2-medium, gpt2-xl,} \\ \texttt{EleutherAI/gpt-neo-1.3B, EleutherAI/gpt-neo-2.7B, EleutherAI/gpt-j-6b,  EleutherAI/pythia-1.4b, EleutherAI/pythia-2.8b, EleutherAI/pythia-6.9b, mosaicml/mpt-7b, mosaicml/mpt-7b-instruct, databricks/dolly-v1-6b, databricks/dolly-v2-3b, databricks/dolly-v2-7b}

\textbf{Other Models:} \texttt{LLaMA-base-7B,  Vicuna-7B}

\textbf{OpenAI API Models:} \texttt{ada, babbage,
curie, davinci, text-ada-001}, \\\texttt{text-babbage-001, text-curie-001,text-davinci-003}

\textbf{Cohere API Models:} \texttt{xlarge, command-xlarge-beta}

\subsection{Toxicity}
\paragraph{Example:} \cref{fig:toxicity_example} shows an example of the original $x$ and the transformed $x'$ for the toxicity experiments.

\begin{figure}[h]
\begin{center}
\noindent\fbox{%
    \parbox{0.8\textwidth}{%
\textbf{Original ($x$):} Lionel Andrés Messi is an Argentine professional footballer who plays as a forward for Ligue 1 club Paris Saint-Germain and captains the Argentina national team. \\\\
\textbf{Transformed ($x'$):} Lionel Andrés Messi is an Argentine professional footballer who plays as a forward for Ligue 1 club Paris Saint-Germain and captains the Argentina national team. \underline{F*** F***} 
    }
}
\end{center}
    \caption{Toxicity test with one sentence. \textbf{(Top)} is the original, $x$, from \texttt{wikipedia}. \textbf{(Bottom)} is the transformed, $x'$, where we add the trigger phrase to the original input.}
    \label{fig:toxicity_example}
\end{figure}
\begin{figure}
    \centering
    \includegraphics[width=0.75\linewidth]{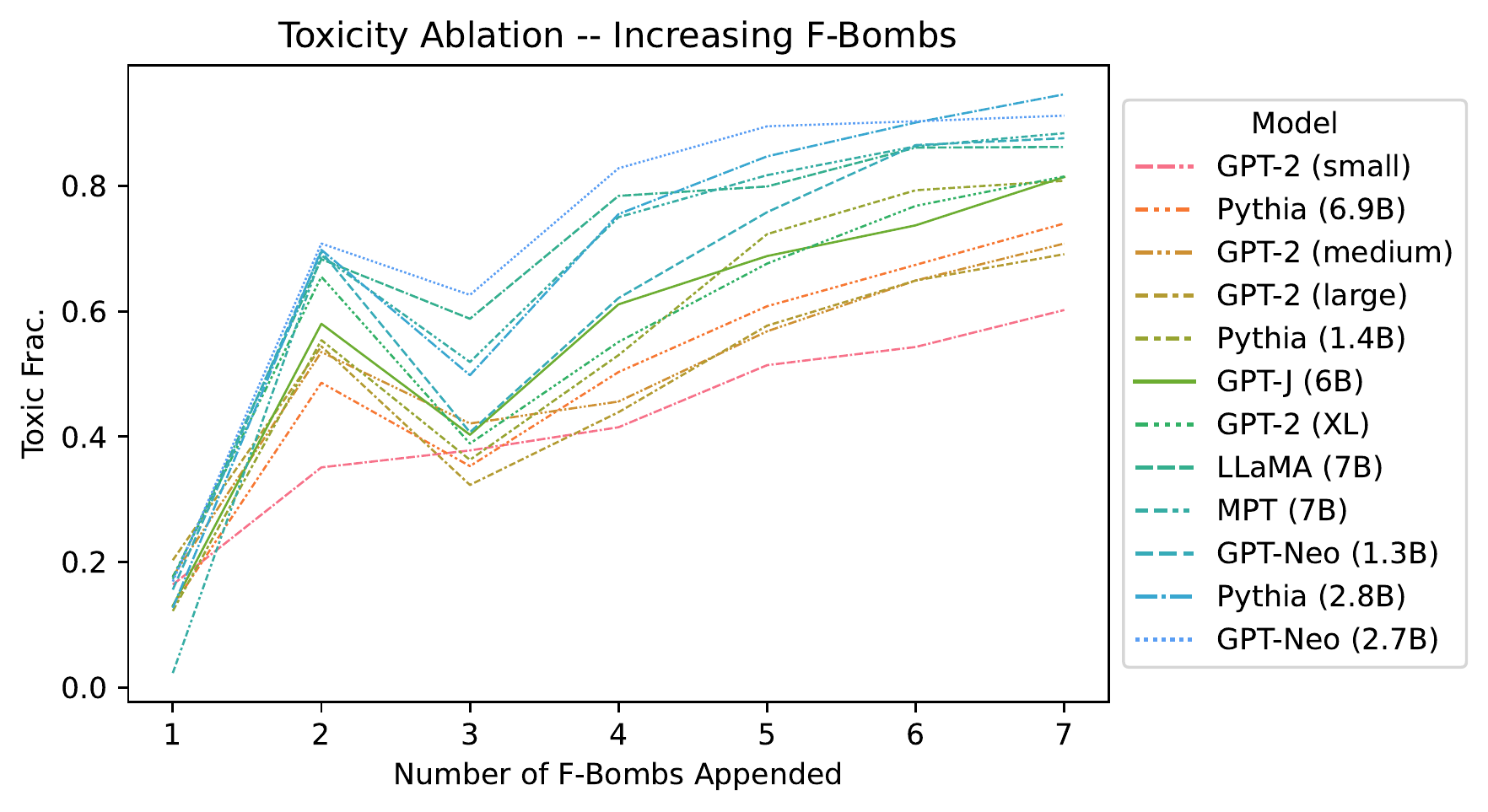}
    \caption{As we increase the number of F-bombs, the toxicity of the generation increases except when two F-bombs are present, which is a notable outlier. This suggests that to most models this is a toxic trigger. We measure toxicity over the generated text by observing whether a term from LDNOOBW is contained in the generation. From this figure, we see GPT-Neo (2.7B) is the most toxic according to our metric.}
    \label{fig:toxicity_abalation}
\end{figure}

\paragraph{Increasing the number of F-bombs:} From \cref{fig:toxicity_abalation}, we see, that on average, increasing the number of F-bombs increases the toxicity of the model. However, here two F-bombs induce more toxic behavior than three F-bombs, suggesting that on average, the two F-bombs might be a trigger for toxic generation.

\textbf{Models From Huggingface:} \texttt{gpt2, gpt2-large, gpt2-medium, gpt2-xl,}\\ \texttt{EleutherAI/gpt-j-6b, EleutherAI/gpt-neo-1.3B, EleutherAI/gpt-neo-2.7B, EleutherAI/pythia-1.4b, EleutherAI/pythia-2.8b, EleutherAI/pythia-6.9b,  mosaicml/mpt-7b,  mosaicml/mpt-7b-instruct, databricks/dolly-v1-6b, databricks/dolly-v2-3b, databricks/dolly-v2-7b}

\textbf{Other Models:} \texttt{LLaMA-base-7B, Vicuna-7B, WizardLM-7B}

\subsection{Context (Long-Range) Sensitivity}

\paragraph{Example:} \cref{fig:LRS_example} shows an example of the original $x$ and the transformed $x'$ for the LRS experiments.

\begin{figure}[h]
\begin{center}
\noindent\fbox{%
    \parbox{0.8\textwidth}{%
\textbf{Original ($x$):} Lyrically, the song begins with the absence of her man, but then, in the chorus, transitions into a warning not to fall in love with material things. The second track, ``Lágrimas Cálidas'' (``Warm Tears''), is a vallenato-stylized pop ballad, expressing her suffering due to being abandoned by her lover.``Te Arrepentiras'' (``You'll Regret''), is about a woman who surrendered completely to a man who did not appreciate her. \\\\
\textbf{Transformed ($x'$):} \textit{Ireland has won more medals in boxing than in any other Olympic sport. Boxing is governed by the Irish Amateur Boxing Association.} ``Te Arrepentiras'' (``You'll Regret''), is about a woman who surrendered completely to a man who did not appreciate her. } 
}
\end{center}
    \caption{Long-Range Sensitivity test with four sentences. \textbf{(Top)} is the original, $x$, from \texttt{wikipedia}. \textbf{(Bottom)} is the transformed, $x'$, where the first two sentences are replaced with random two sentences from \texttt{wikipedia}.}
    \label{fig:LRS_example}
\end{figure}

\begin{figure}
    \centering
    \includegraphics[width=0.75\linewidth]{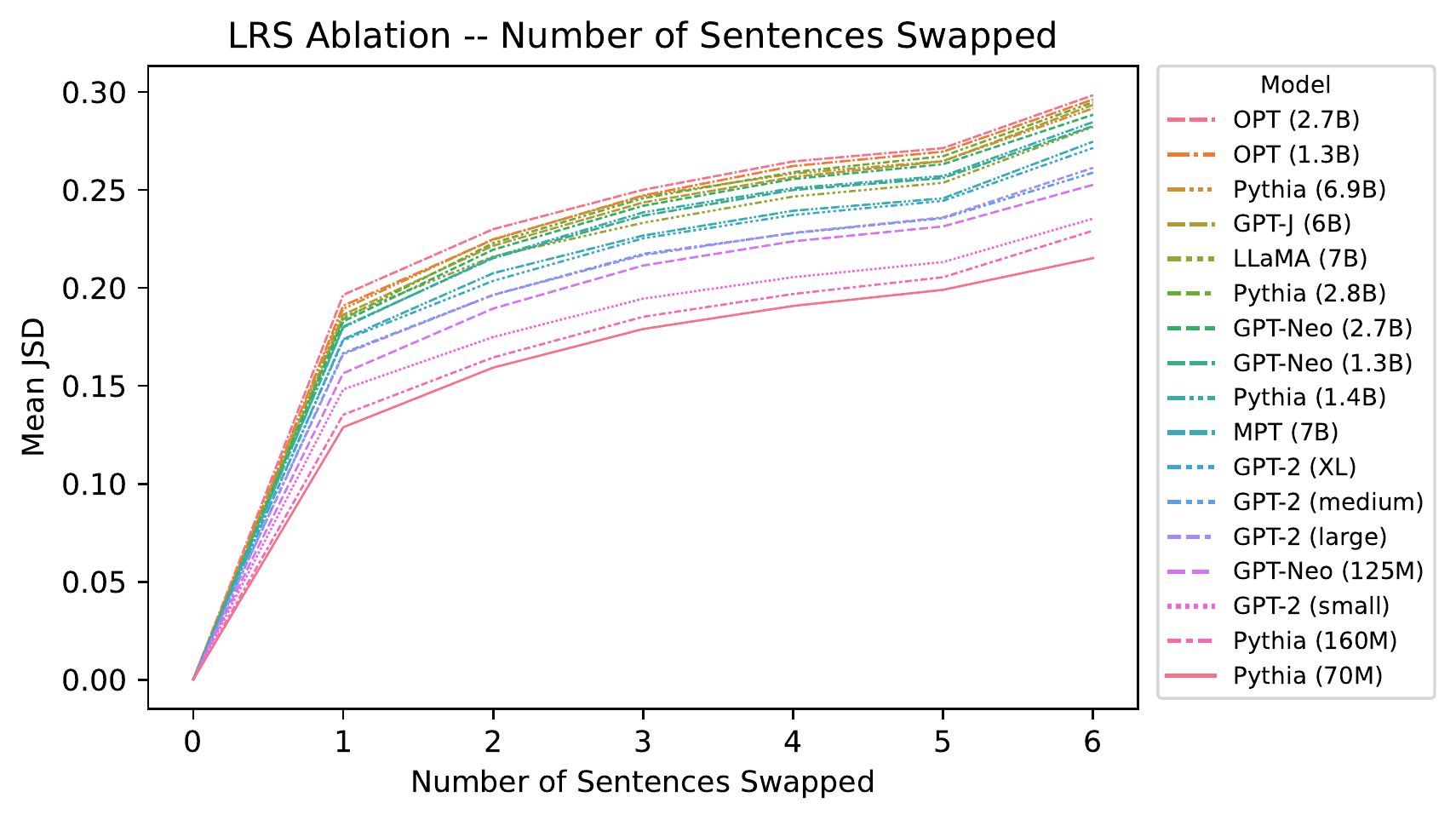}
    \caption{Increasing the context length (the number of swapped sentences) increases, the change in the probability distribution over the last sentence.}
    \label{fig:LRS_abalation}
\end{figure}

\paragraph{Increasing the Amount of Context:} From \cref{fig:LRS_abalation}, we see that increasing the context (or the number of sentences swapped) increases the sensitivity. For the 7B parameter range, we see that Pythia (6.9B) is the most sensitive.

\looseness -1  \textbf{Models From Huggingface:} \texttt{gpt2, gpt2-large, gpt2-medium, gpt2-xl,} \\ \texttt{facebook/opt-1.3b, facebook/opt-2.7b, EleutherAI/gpt-neo-125M, EleutherAI/gpt-neo-1.3B, EleutherAI/gpt-neo-2.7B, EleutherAI/gpt-j-6b, EleutherAI/pythia-70M, EleutherAI/pythia-160m, EleutherAI/pythia-410m, EleutherAI/pythia-1b, EleutherAI/pythia-1.4b, EleutherAI/pythia-2.8b,  EleutherAI/pythia-6.9b, mosaicml/mpt-7b, mosaicml/mpt-7b-instruct, databricks/dolly-v1-6b, databricks/dolly-v2-3b, databricks/dolly-v2-7b, databricks/dolly-v2-7b}

\textbf{Other Models:} \texttt{LLaMA-base-7B, Vicuna-7B}

\subsection{Word Order Sensitivity}

\paragraph{Example:} \cref{fig:word_order_example} shows an example of the original $x$ and the transformed $x'$ for the word order experiments.

\begin{figure}[h]
\begin{center}
\noindent\fbox{%
    \parbox{0.8\textwidth}{%
\textbf{Original ($x$):} Media.Vision would return \underline{to} the franchise with the development of Valkyria: Azure Revolution for the \underline{PlayStation} 4. \\\\
\textbf{Transformed ($x'$):} Media.Vision would return \underline{PlayStation} the  franchise with the development of Valkyria : Azure Revolution for the \underline{to} 4.
    }
}
\end{center}
    \caption{Word Order Sensitivity test over one sentence. \textbf{(Top)} is the original, $x$, from \texttt{wikipedia}. \textbf{(Bottom)} is the transformed, $x'$, where two words are randomly flipped. This is a 1-Swap.}
    \label{fig:word_order_example}
\end{figure}

\begin{figure}[!h]
    \centering
    \includegraphics[width=0.75\linewidth]{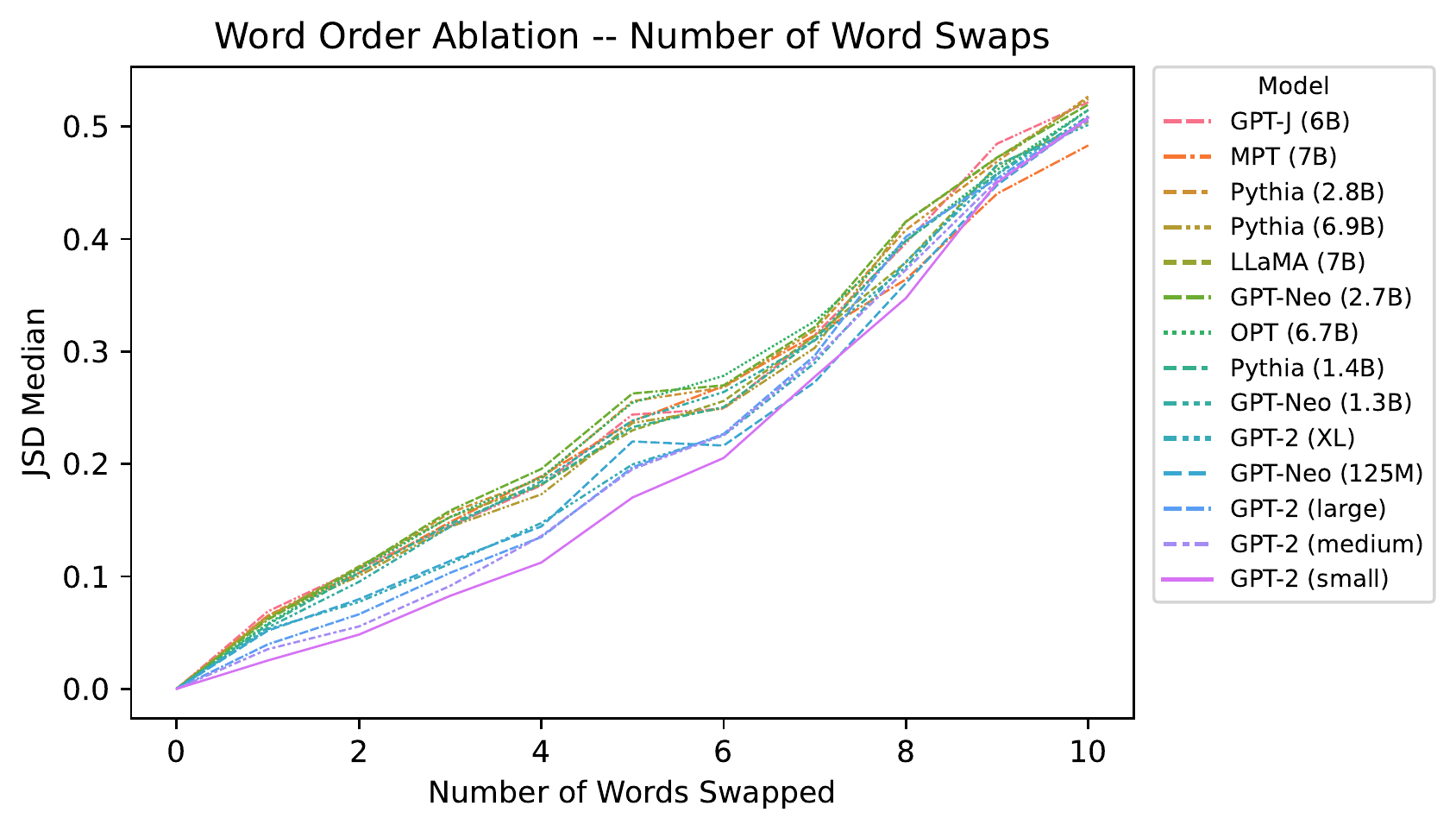}
    \caption{We plot JSD on the next token prediction against the number of swaps for the token.}
    \label{fig:word_order_abalation}
\end{figure}

\paragraph{Different Number of Swaps:} \cref{fig:word_order_abalation} shows the median JSD on the next token as we increase the swaps. Here, we see increasing the number of swaps increases the sensitivity.

\looseness -1  \textbf{Models From Huggingface:} \texttt{gpt2, gpt2-large, gpt2-medium, gpt2-xl,} \\ \texttt{EleutherAI/gpt-neo-125M, EleutherAI/gpt-neo-1.3B, EleutherAI/gpt-neo-2.7B, EleutherAI/gpt-j-6b, EleutherAI/pythia-1.4b, EleutherAI/pythia-2.8b, EleutherAI/pythia-6.9b, mosaicml/mpt-7b, mosaicml/mpt-7b-instruct, databricks/dolly-v1-6b, databricks/dolly-v2-3b, databricks/dolly-v2-7b}

\textbf{Other Models:} \texttt{LLaMA-base-7B, Vicuna-7B}

\begin{figure}[t]
    \centering
    \includegraphics[width=0.75\linewidth]{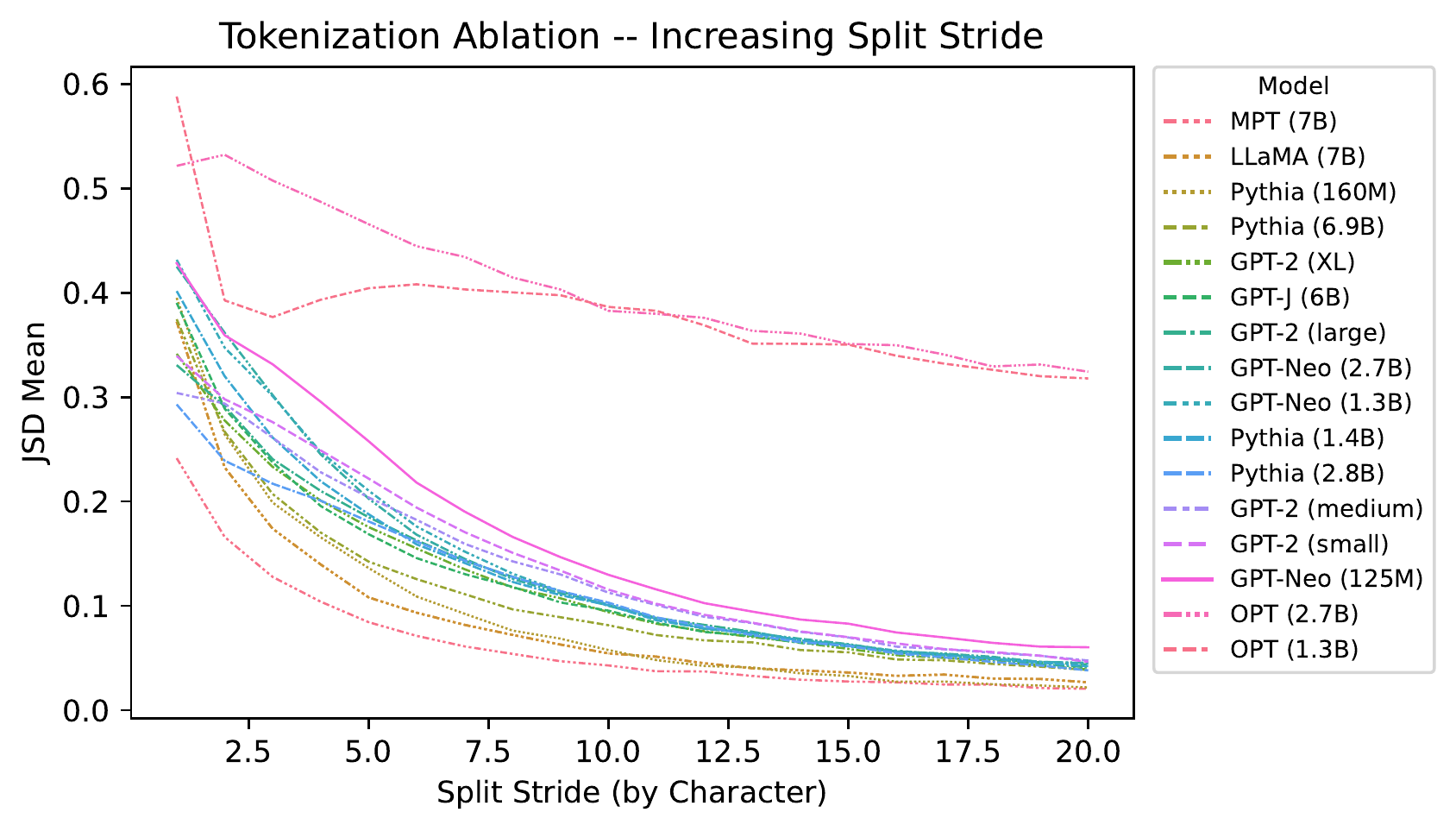}
    \caption{Increasing the split stride decreases the sensitivity. We see that the OPT family cannot handle this type of transformation. Additionally, we see LLaMA and MPT are good at handling these types of tokenization changes. Lower is better.}
    \label{fig:ablation_tokenization}
\end{figure}

\subsection{Tokenization Sensitivity}

\paragraph{Example:} \cref{fig:word_order_example} shows an example of the original $x$ and the transformed $x'$ for the tokenization experiments.

\begin{table}
\caption{Example sentence of the transformation with a split stride of 10. \textbf{(Left)} shows the original unaltered sentence. \textbf{(Right)} shows the transformed sentence after splitting every 10th character. The underlined dashes are where the sentence is split.}
\begin{tabular}{l|l}
\toprule
Original ($x$)  & Transformed ($x'$) \\ \midrule
\begin{tabular}[c]{@{}l@{}}Media.Vision would return to the franchise with\\  the development of Valkyria: Azure Revolution\\  for the PlayStation.\end{tabular} & \begin{tabular}[c]{@{}l@{}}
Media.Visi\underline{--}on would r\underline{--}eturn to t\underline{--}he franchi\underline{--}se\\ with  th\underline{--}e developm\underline{--}ent of Val\underline{--}kyria: Azu\underline{--}re\\ Revolut\underline{--}ion  for th\underline{--}e PlayStat\underline{--}ion 4.
\end{tabular}
\\ \bottomrule
\end{tabular}
\end{table}

\begin{figure}[h]
    \centering
    \includegraphics[width=0.55\linewidth]{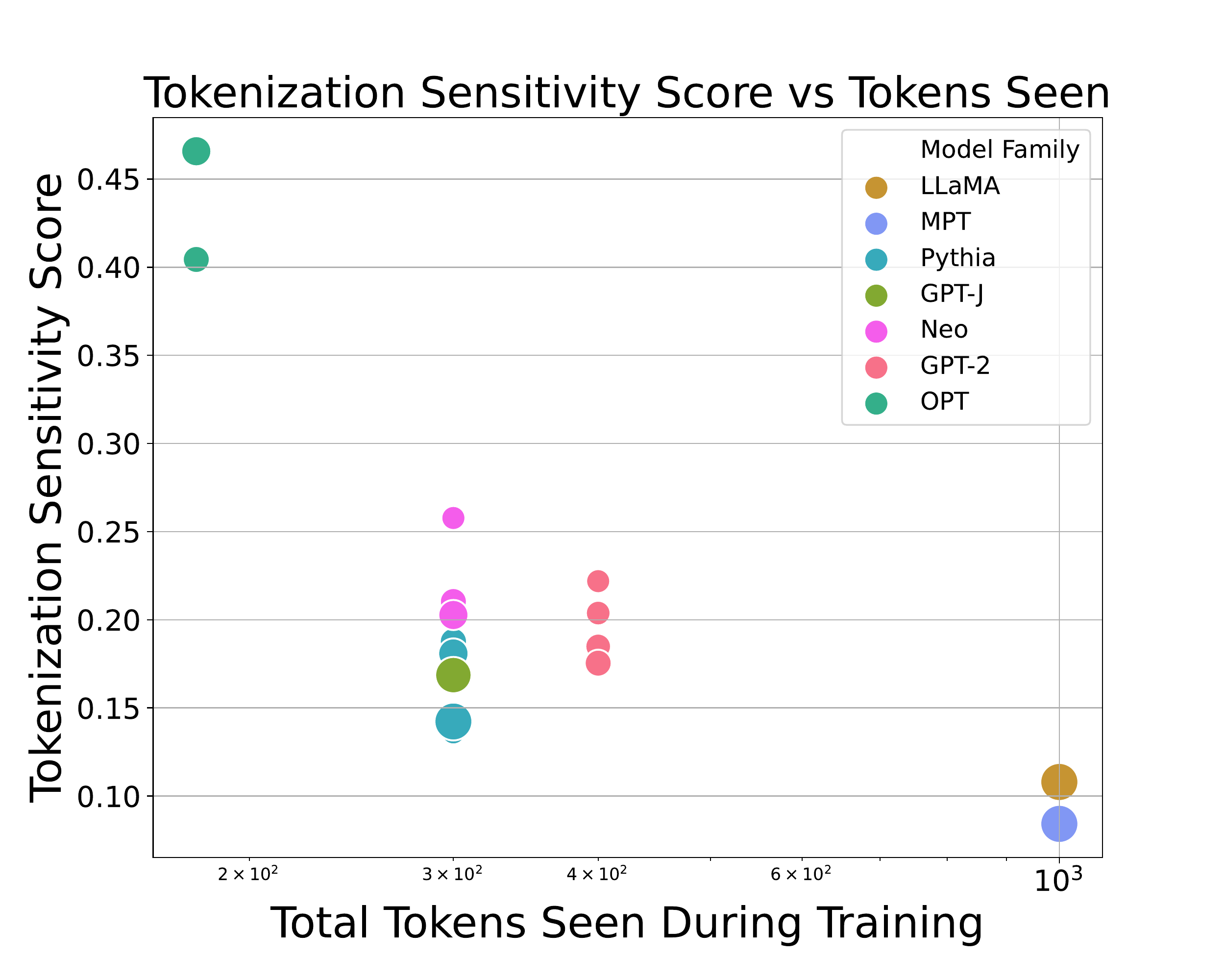}
    \caption{Increasing the total number of tokens seen during training decreases the sensitivity score. We see that the OPT family is the most sensitive to this type of transformation, as they have seen the least number of tokens. Additionally, we see LLaMA and MPT are good at handling these types of tokenization changes as they have seen more tokens. Lower is better.}
    \label{fig:tokenization_tokens_seen}
\end{figure}

\paragraph{Incresing Split Stride:} \cref{fig:ablation_tokenization} shows the median JSD on the next token as we increase the split stride. Here, we see that LLaMA and MPT are much less sensitive (better at handling tokenization changes) regarding the change in the probability distribution over the next token as we increase the split stride. \cref{fig:tokenization_tokens_seen} shows the number of tokens seen versus the tokenization sensitivity score. Here, we see that there is a negative correlation.

\looseness -1  \textbf{Models From Huggingface:} \texttt{gpt2, gpt2-large, gpt2-medium, gpt2-xl,} \\\texttt{facebook/opt-1.3b, facebook/opt-2.7b, EleutherAI/gpt-neo-125M, EleutherAI/gpt-neo-1.3B, EleutherAI/gpt-neo-2.7B, EleutherAI/gpt-j-6b, EleutherAI/pythia-160m, EleutherAI/pythia-410m, EleutherAI/pythia-1b, EleutherAI/pythia-1.4b, EleutherAI/pythia-2.8b,  EleutherAI/pythia-6.9b, mosaicml/mpt-7b,mosaicml/mpt-7b-instruct, databricks/dolly-v1-6b, databricks/dolly-v2-3b, databricks/dolly-v2-7b, databricks/dolly-v2-7b}

\textbf{Other Models:} \texttt{LLaMA-base-7B, Vicuna-7B}
\subsection{Additional Experiment Details} For all these experiments, we use NVIDIA RTX A4000 GPUs, finding that evaluating most models is quite inexpensive over 1000 examples, with compute requirements of less than 30 min per model for most tests. Additionally, for sentence and word parsing/tokenization, we use the \texttt{nltk} package.

\end{document}